\documentclass{article}
\PassOptionsToPackage{square,comma,numbers,compress}{natbib}

\usepackage[preprint]{neurips_2023}

\usepackage[utf8]{inputenc} %
\usepackage[T1]{fontenc}    %
\usepackage[hidelinks,colorlinks=true]{hyperref}       %
\usepackage[dvipsnames]{xcolor}
\usepackage{url}            %
\usepackage{booktabs}       %
\usepackage{amsfonts}       %
\usepackage{nicefrac}       %
\usepackage{microtype}      %
\usepackage{xcolor}         %
\usepackage{titlesec}
\usepackage{graphicx}
\usepackage{multirow}
\usepackage{amsmath}
\usepackage{hyperref}
\graphicspath{ {figures/} }
\usepackage{array}

\usepackage{pgfplots}
\pgfplotsset{compat=1.7}

\usepackage{times,graphicx,tabulary,multirow,xspace}
\usepackage{subfig}
\usepackage{changepage}

\title{A Preliminary Exploration with GPT-4o Voice Mode}

\author{
\textbf{Yu-Xiang Lin}\quad \textbf{Chih-Kai Yang}\quad \textbf{Wei-Chih Chen}\quad \textbf{Chen-An Li} \\
\textbf{Chien-yu Huang} \quad \textbf{Xuanjun Chen} \quad \textbf{Hung-yi Lee} \\ \\
National Taiwan University
}

\begin{document}

\maketitle

\begin{abstract}
With the rise of multimodal large language models, GPT-4o\footnote{This report referred to the model with audio processing capability as ``GPT-4o'', following the OpenAI report~\cite{hurst2024gpt}. We used the ``gpt-4o-audio-preview-2024-10-01'' version to conduct the following experiments.} stands out as a pioneering model, driving us to evaluate its capabilities. This report assesses GPT-4o  across various tasks to analyze its audio processing and reasoning abilities. We find that GPT-4o exhibits strong knowledge in audio, speech, and music understanding, performing well in tasks like intent classification, spoken command classification, semantic and grammatical reasoning., multilingual speech recognition, and singing analysis. It also shows greater robustness against hallucinations than other large audio-language models (LALMs). However, it struggles with tasks such as audio duration prediction and instrument classification. Additionally, GPT-4o's safety mechanisms cause it to decline tasks like speaker identification, age classification, MOS prediction, and audio deepfake detection. Notably, the model exhibits a significantly different refusal rate when responding to speaker verification tasks on different datasets. This is likely due to variations in the accompanying instructions or the quality of the input audio, suggesting the sensitivity of its built-in safeguards. Finally, we acknowledge that model performance varies with evaluation protocols. This report only serves as a preliminary exploration of the current state of LALMs.

\end{abstract}

\section{Introduction}

Recent advances in spoken dialogue systems, such as Gemini-1.5~\cite{team2024gemini}, GPT-4o~\cite{hurst2024gpt}, Moshi-Chat~\cite{defossez2024moshi} have captured significant attention. Unlike cascaded pipelines that integrate a speech recognition model with a text-based large language model (LLM), end-to-end LALMs excel at capturing rich information embedded in audio inputs that are often absent in ASR transcriptions, such as prosody, emotion, speaker information, and environmental sounds. The lack of such capabilities propels the development of end-to-end LALMs, as these acoustic features are valuable for a voice assistant. Besides, cascaded systems suffer from error propagation, further highlighting the advantages of end-to-end approaches. Building upon these viewpoints,  this report aims to address the question: ~\textit{How close are we to achieving a universal instruction-based speech model?} Among the various LALMs, GPT-4o emerges as one of the most highly anticipated models, driving us to evaluate its performance to explore this question in depth.

To thoroughly assess the audio understanding and reasoning capabilities of GPT-4o, we conduct comprehensive experiments on a wide range of tasks, analyzing the model's performance across various criteria. This report includes evaluations on massive benchmarks including Dynamic-SUPERB Phase2~\cite{huang2024dynamic}\footnote{There is a preceding work named Dynamic-SUPERB~\cite{huang2024dynamic1}, which can be viewed as a subset of Phase 2. For simplicity, we refer to Dynamic-SUPERB Phase 2 as Dynamic-SUPERB in the following sections.}, MMAU~\cite{sakshi2024mmau} and CMM~\cite{leng2024curse}, spanning the domains of~\textbf{Audio},~\textbf{Speech}, and~\textbf{Music}.  These benchmarks enable us to evaluate the effectiveness of GPT-4o in interpreting acoustic information from audio inputs and responding based on relevant observations. 

Specifically, Dynamic-SUPERB is a large-scale benchmark designed for evaluating instruction-based universal speech models, covering hundreds of diverse tasks. In contrast, MMAU is designed for assessing the reasoning and understanding abilities of LALMs, while CMM focuses on measuring the hallucination levels in LMMs. Interestingly, cascaded pipelines and random guessing sometimes outperform or match the performance of end-to-end LALMs. For tasks heavily reliant on acoustic information, cascaded pipelines can achieve better results by relying solely on text transcriptions. Similarly, random guessing in classification tasks occasionally outperforms LALMs, highlighting weaknesses in their ability to follow instructions effectively. These findings indicate the direction of development for future LALMs.

Additionally, we find that the safeguard mechanism built into the current GPT-4o is highly sensitive, as it refuses to process certain tasks that do not involve safety concerns, such as speaker count identification. Furthermore, GPT-4o demonstrates inconsistent behavior across different samples of the same task, showing a higher refusal rate on some datasets while being more permissive on others. We speculate that this variation stems from differences in the accompanying instructions or the quality of the input audio, which affect the difficulty level of the samples. Another possible reason is the post-training effect, which makes GPT-4o avoid hallucinations and prevent misleading users when processing difficult samples.

Regardless of the cause, the safeguard mechanism poses a challenge in accurately assessing the true capabilities of proprietary models like GPT-4o. However, to ensure a fair comparison with other LALMs, we do not employ alternative methods to override or bypass these safeguards. We emphasize that model performance may vary with evaluation protocols. This report only serves as a preliminary exploration of the current state of LALMs.

\section{Development of Large Audio-Language Models}

Recent works on developing LALMs can be roughly divided into two parallel lines of work. One leverages strong text-based LLMs as the backbone and incorporates specialized models for speech, audio, and music processing to build cascaded systems capable of dealing with diverse tasks from these domains. In contrast, the other aims to build end-to-end language models having the multi-modality understanding to directly solve the tasks of these modalities without relying on specialized models. In this section, we briefly summarize the current status of these two lines of work.

We first introduce the cascaded systems. AudioGPT~\cite{huang2023audiogpt} proposed to use LLM as the controller, which is responsible for analyzing the user query, assigning one suitable speech/audio/music model to solve the encountered task, and sending the results as a response to the user. Speech-Copilot~\cite{kuan2024speech} further proposed to formulate the specialized models from the audio/speech/music domains to be callable tools and allow the LLMs to integrate these tools via programming, which further enhanced the flexibility of the combination of the specialized models. These works have demonstrated the effectiveness of using the strong reasoning ability of LLMs as the basis of audio/speech/music processing applications. 

On the other hand, there are also several efforts to develop end-to-end language models possessing multi-modality understanding. dGSLM~\cite{nguyen2023generative} serves as the first end-to-end speech LLM trained with two-channel raw conversational audio, without any text or label involved. It integrates a HuBERT-based quantizer, a dual-tower transformer, and a multi-speaker HiFi-GAN~\cite{kong2020hifi} vocoder . Building on a similar architecture, SpeechGPT~\cite{zhang2023speechgpt} replaces the language model backbone with LLaMA~\cite{touvron2023llama} to enhance performance by utilizing the transfer capabilities of textual LLMs. Further advancements have been made with models like Qwen-Audio~\cite{chu2023qwen}, SALMONN~\cite{tang2024salmonn}, and Qwen-2-Audio~\cite{chu2024qwen2}, which incorporate improved architectural designs and leverage larger datasets for scaling and enhanced functionality. Similarly, MU-LLaMA~\cite{liu2024music} combines a pre-trained MERT model and LLaMA-2~\cite{touvron2023llama2}, while its focus is music-related tasks like music question answering and music caption generation. Taking an alternative approach, WavLLM~\cite{hu2024wavllm} adopts a two-stage curriculum learning framework, progressing from basic single tasks to advanced multi-task optimization. GAMA-IT~\cite{deshmukh2024audio}, on the other hand, introduces a multilayer aggregator module to capture multi-scale audio information.

Additionally, to enhance speech-instruction-following capabilities and enable streaming interactions, Mini-Omni~\cite{xie2024mini} and LLaMA-Omni~\cite{fang2024llama} construct datasets comprising speech instructions and corresponding responses by rephrasing text instruction data and synthesizing speech data. However, their training datasets are limited to speech data, excluding audio and music data, which limits their ability to process these types of inputs effectively. 

To align language model with human preference, reinforcement learning from human feedback (RLHF)~\cite{ouyang2022training} or reinforcement learning from AI feedback (RLAIF)~\cite{10.5555/3692070.3693141} have been widely adopted in developing textual LLMs like GPT-4~\cite{hurst2024gpt} or LLaMA-2. SpeechAlign~\cite{zhang2024speechalign} is the first work to enhance the LALMs by mitigating the misalignment between training and testing inputs through learning from human feedback. Align-SLM~\cite{lin2024align} constructs and learns from a semantic preference dataset by sampling multiple speech completions, transcribing and scoring them by a Mistral model~\cite{jiang2023mistral}.  Beyond improving performance, RL-based post-training is commonly used to enhance the safety and reliability of LLMs. For example, Qwen-2-Audio, LLaMA-3~\cite{dubey2024llama} and Gemini-1.5~\cite{team2024gemini} incorporate these techniques to further improve their alignment and safeguard against undesirable outputs.

Among these LALMs, GPT-4o possesses the aforementioned capabilities and is widely anticipated by researchers, serving as the representative model of current state of LALMs. Therefore, this report focuses on analyzing its capabilities to address the posed questions.

\section{Overview}
\subsection{Comprehensive Evaluation}
We conduct experiments on a variety of benchmarks to comprehensively analyze GPT-4o. For a holistic assessment, we use Dynamic-SUPERB~\cite{huang2024dynamic}, which evaluates models across hundreds of diverse tasks, allowing us to gauge GPT-4o’s holistic capabilities. Additionally, we employ specialized benchmarks to assess specific skills such as ~\textbf{advanced domain knowledge},~\textbf{acoustic information reasoning}, and~\textbf{ability to identify hallucination}. For tasks requiring expert-level knowledge and audio reasoning capabilities, we use MMAU\citep{sakshi2024mmau}; for hallucination detection, CMM\cite{leng2024curse} is adopted. Note that the results of baseline models are taken from the original paper publishing the benchmark, and the credit should go to the original paper. As all the above benchmarks provide official evaluation systems or scripts, it's reasonable that the results between GPT-4o and other baseline models are fairly comparable since the following results are finished with the given pipelines.

\subsection{Refusal Detection}
Besides, we observe that GPT-4o performs nearly no correct prediction on some tasks in Dynamic SUPERB, on which other baseline models perform well. Based on manual inspection, we find that GPT-4o tends to respond with sentences with the same pattern like ``Sorry, I'm not able to perform such task'' or ``I'm sorry, I can't analyze or classify the musical genre of audio files. Can I help you with anything else?''. According to the OpenAI report~\cite{hurst2024gpt}, GPT-4o undergoes additional post-training procedures designed to reduce risks and integrates specialized classifiers to block or rephrase specific types of outputs in the deployed system. The key categories of restricted outputs are as follows:

\begin{itemize}
    \item \textbf{Violative and Disallowed Content} includes content that is illegal, such as explicit erotic material, depictions of violence, self-harm, or other harmful content.
    \item \textbf{Misinformation} refers to inaccurate information that could deceive or mislead users.
    \item \textbf{Generation of Copyrighted Material} covers unauthorized reproduction of copyrighted music, audio, or speech associated with specific speakers.
    \item \textbf{Ungrounded Inferences} refers to queries about a speaker that make unfounded or unsupported claims without any reliable sources, including aspects such as the speaker's race, socioeconomic status or occupation, religious beliefs, personality traits, political attributes, appearance, gender, and more.
    \item \textbf{Attribution of Sensitive Traits} refers to queries about personal traits, such as a speaker's accent or nationality. Unlike Ungrounded Inferences, this type of inference could feasibly be determined based solely on audio.
    \item \textbf{Prohibited Audio Content} includes outputs related to restricted or banned audio types.
    \item \textbf{Speaker Identification}: Identifies who is speaking.
    \item \textbf{Outputs Containing Erotic or Violent Speech} Involves responses that include sexually explicit or violent language.
\end{itemize}

As stated in the official report, GPT-4o was post-trained to refuse all the aforementioned requests except for~\textbf{Attribution of Sensitive Traits}, where a hedging strategy is applied to responses. To evaluate the effectiveness of this post-training, we report the refusal rate for each task in the following section. To efficiently identify the refusal produced by the model, we employ two refusal detection strategies. The first is a string-matching method, as we observe that GPT-4o often refuses by template-based responses, which include keywords such as ``sorry'' or ``apologize''. The second involves using LLM as a judge. Specifically, we use LLaMA-3.1-8B-Instruct~\cite{dubey2024llama} to determine whether GPT-4o's response for each task sample constitutes a refusal. The instruction adopted here is~\textit{``Given the Instruction/Question, does the model's response indicate a refusal to perform the task/answer the question? Respond with Yes or No.''} We report the value determined by the string matching method as ``Refusal Rate-Str'' and the value determined by the LLM judge as ``Refusal Rate-LLM''.

\section{Dynamic-SUPERB}
\label{sec:dynamic-superb}
Dynamic-SUPERB~\cite{huang2024dynamic} serves as a benchmark designed to assess the understanding and instruction-following capabilities of voice models. This benchmark comprises 180 tasks, contributed by the global research community, and builds upon existing benchmarks. Specifically, it defines its core tasks by reformulating three established benchmarks: SUPERB~\cite{yang21c_interspeech}, HEAR~\cite{turian2022hear}, and MARBLE~\cite{yuan2023marble}, encompassing tasks related to speech, audio, and music. 

Referencing sessions from the INTERSPEECH conference and EDICS of IEEE SPS, Dynamic-SUPERB develops a task taxonomy to interpret performance results across various tasks. They categorize tasks into 17 domains:

\vspace{0.5cm}
\noindent
\begin{tabular}{@{\hspace{-0.9cm}}p{0.45\textwidth} @{\hspace{-0.9cm}}p{0.3\textwidth} @{\hspace{-0.4cm}}p{0.45\textwidth}}

    \hspace{75px}\textbf{Speech} & \hspace{55px}\textbf{Music} & \hspace{65px}\textbf{Audio} \\
    \begin{itemize}
        \item Paralinguistics
        \item Phonetics, Phonology, Prosody
        \item Safety and Security
        \item Speaker \& Language
        \item Speech Enhancement
        \item Speech Recognition
        \item Speech, Voice, Hearing Disorder
        \item Spoken Language Understanding
    \end{itemize} &
    \begin{itemize}
        \item Harmony \& Pitch
        \item Music Classification
        \item Rhythm Analysis
    \end{itemize} &
    \begin{itemize}
        \item Quality Assessment
        \item Safety
        \item Signal-Characteristics Analysis
        \item Singing Analysis
        \item Sound Event
        \item Spatial Audio Analysis
    \end{itemize} \\
\end{tabular}
\vspace{0.2cm}

In this section, we present the evaluation results for each domain, organized according to the taxonomy outlined above. We take the cascaded system Whisper-LLaMA as the baseline model to compute the relative score on each task, following the setting in Dynamic-SUPERB. Additionally, since the relative score becomes meaningless when Whisper-LLaMA performs poorly, we introduce a naive random guess baseline for classification tasks to address this issue. For random baseline, we simply repeats the experiments for 100 times with uniform sampling. That is, for a $n$-class classification problem, given a testing sample, the probability to predict each class is evenly distributed as $\frac{1}{n}$. 

\clearpage
\subsection{Speech Domain - Paralinguistics}

\begin{figure}[h!]
    \centering
    \includegraphics[width=\textwidth]{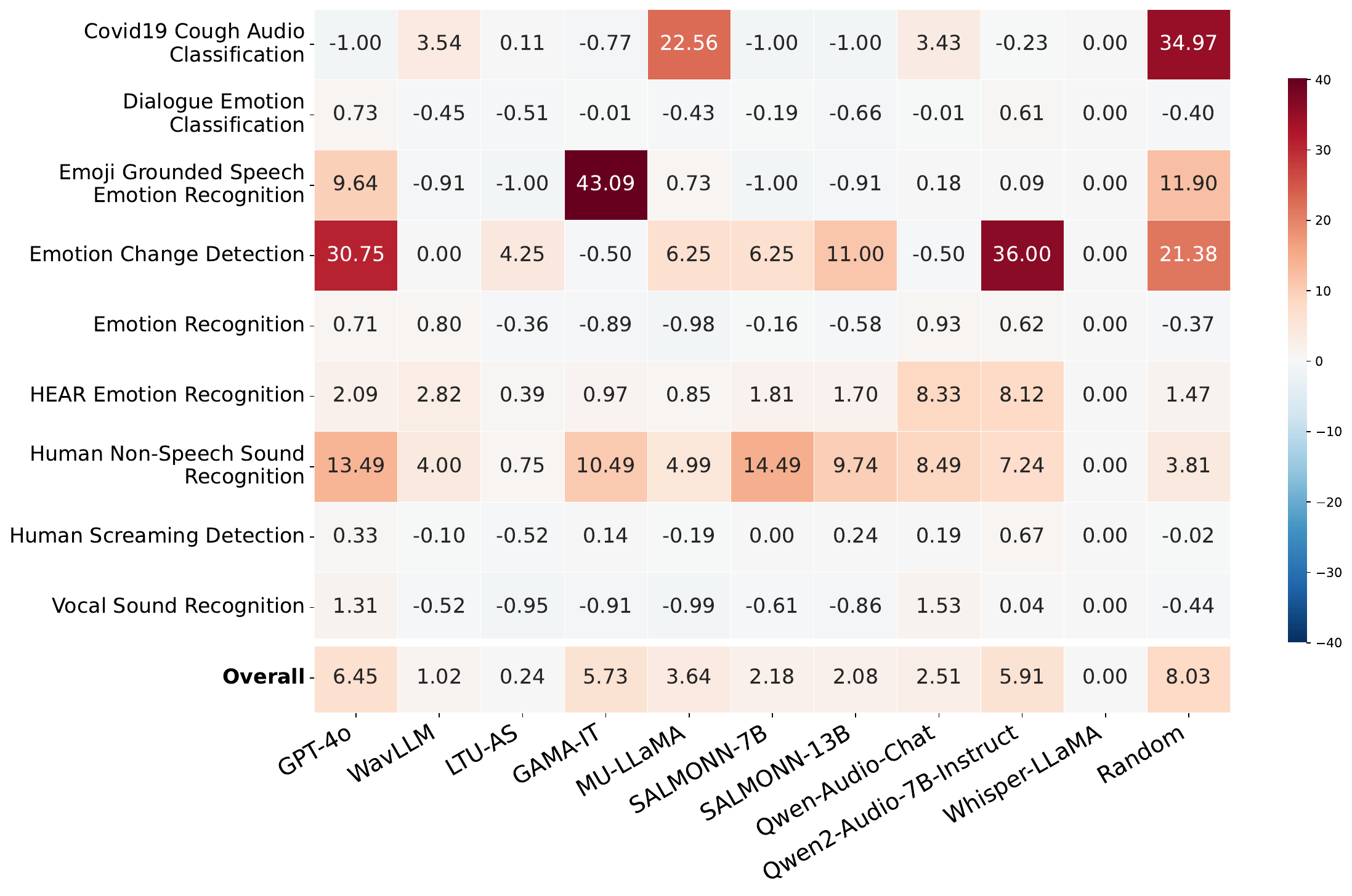} 
    \caption{Relative performance comparison of models in the~\textbf{(Speech) Paralinguistics} domain. }
    \label{fig:heatmap-speech-paralinguistics}
\end{figure}

\begin{figure}[h!]
    \centering
     \includegraphics[width=\textwidth]{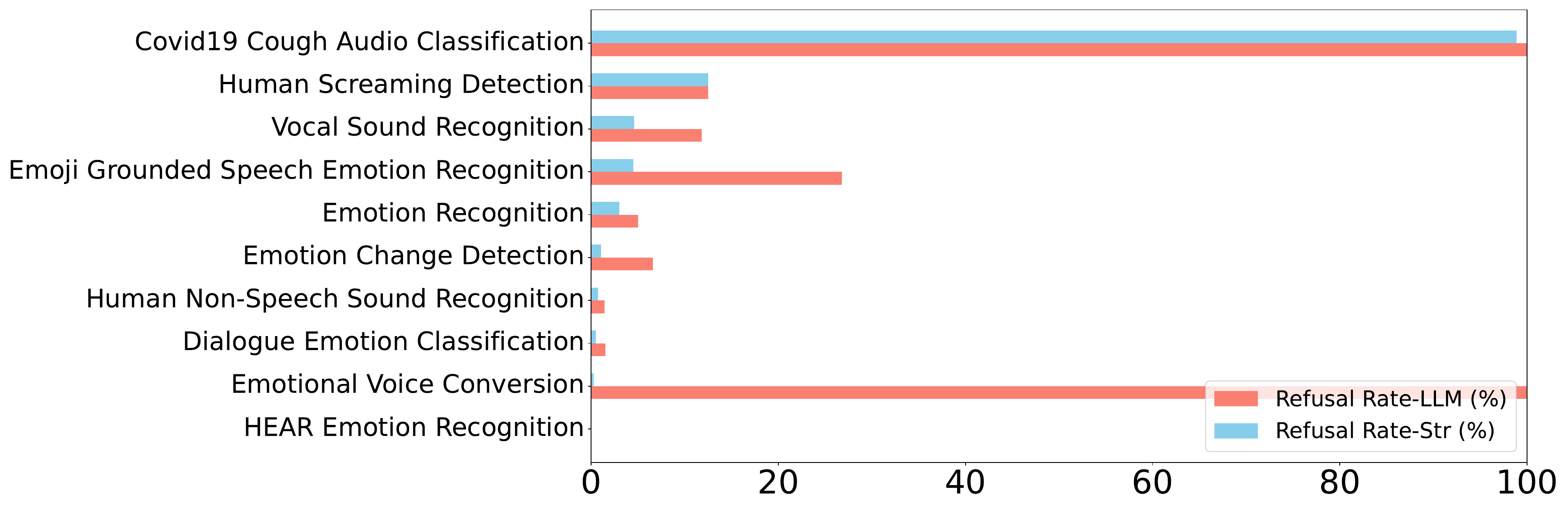} 
    \caption{Refusal rates of tasks in the~\textbf{(Speech) Paralinguistics} domain }
    \label{fig:rejection-speech-paralinguistics}
\end{figure}

The overview, relative scores, and refusal rates of tasks about~\textbf{Paralinguistics} are demonstrated in Table~\ref{tab:tasks-speech-paralinguistics}, Figure~\ref{fig:heatmap-speech-paralinguistics} and Figure~\ref{fig:rejection-speech-paralinguistics} respectively. Based on our conjecture, the primary ethical concern in this domain is the potential for misinformation in the ``Covid19 Cough Audio Classification`` task, as the model could not perform this task accurately. Besides, although the refusal rate of ``Emotional Voice Conversion`` judged by LLM is high, GPT-4o is actually willing to perform the task.~\footnote{As of the release date of this paper, the official evaluation protocol for this task is not yet available, as it is future work under Dynamic SUPERB.} This is because we ask textual LLMs, which process only textual responses, making it difficult to detect rejection on this type of audio generation task accurately.

GPT-4o declines to predict the health condition of individuals in ``Covid19 Cough Audio Classification'', resulting in the worst performance among baseline models. Interestingly, Random baseline outperforms all LALMs and Whisper-LLaMA, reflecting the difficulty current models face in accurately identifying types of coughs. This observation also supports our conjecture that GPT-4o lacks confidence in performing such tasks, leading it to refuse to respond. However, in tasks related to emotion recognition or vocal sound recognition, GPT-4o is comparable to Qwen2-Audio-7B-Instruct and GAMA-IT, outperforming other models, including Whisper-LLaMA. The fact that most LALMs outperform Whisper-LLaMA in most tasks further confirms that LALMs have the upper hand in paralanguage understanding.

\begin{table}[h!]
\centering
\begin{tabular}{p{0.5\textwidth}|p{0.45\textwidth}} 
\toprule
\midrule
\textbf{Task Name} & \textbf{Task Description} \\
\midrule
- Covid19 Cough Audio Classification & Identify the health condition of the person. \\
\midrule
- Dialogue Emotion Classification & \multirow{3}{0.45\textwidth}{Identify the emotion of the speaker} \\
- Emotion Recognition & \\
- HEAR Emotion Recognition &  \\
\midrule
- Emoji Grounded Speech Emotion Recognition & Identify emojis that best represent the speaker's emotional state. \\
\midrule
- Emotion Change Detection & Identify the shifted events of emotions
\\
\midrule 
- Human Non-Speech Sound Recognition & \multirow{3}{0.45\textwidth}{Identify what the type of non-speech sounds produced by speaker is or determine whether it exists}\\
- Human Screaming Detection &  \\
- Vocal Sound Recognition &  \\
\midrule 
- Emotional Voice Conversion & Generate audio by re-speaking the given utterance to convey a specific emotion. \\
\midrule 
\bottomrule
\end{tabular}
\vspace{10px}
\caption{Overview of tasks in the~\textbf{(Speech) Paralinguistics} domain.}
\label{tab:tasks-speech-paralinguistics}
\end{table}

\clearpage


\subsection{Speech Domain - Phonetics, Phonology, Prosody}

\begin{figure}[htbp]
    \centering
    \includegraphics[width=\textwidth]{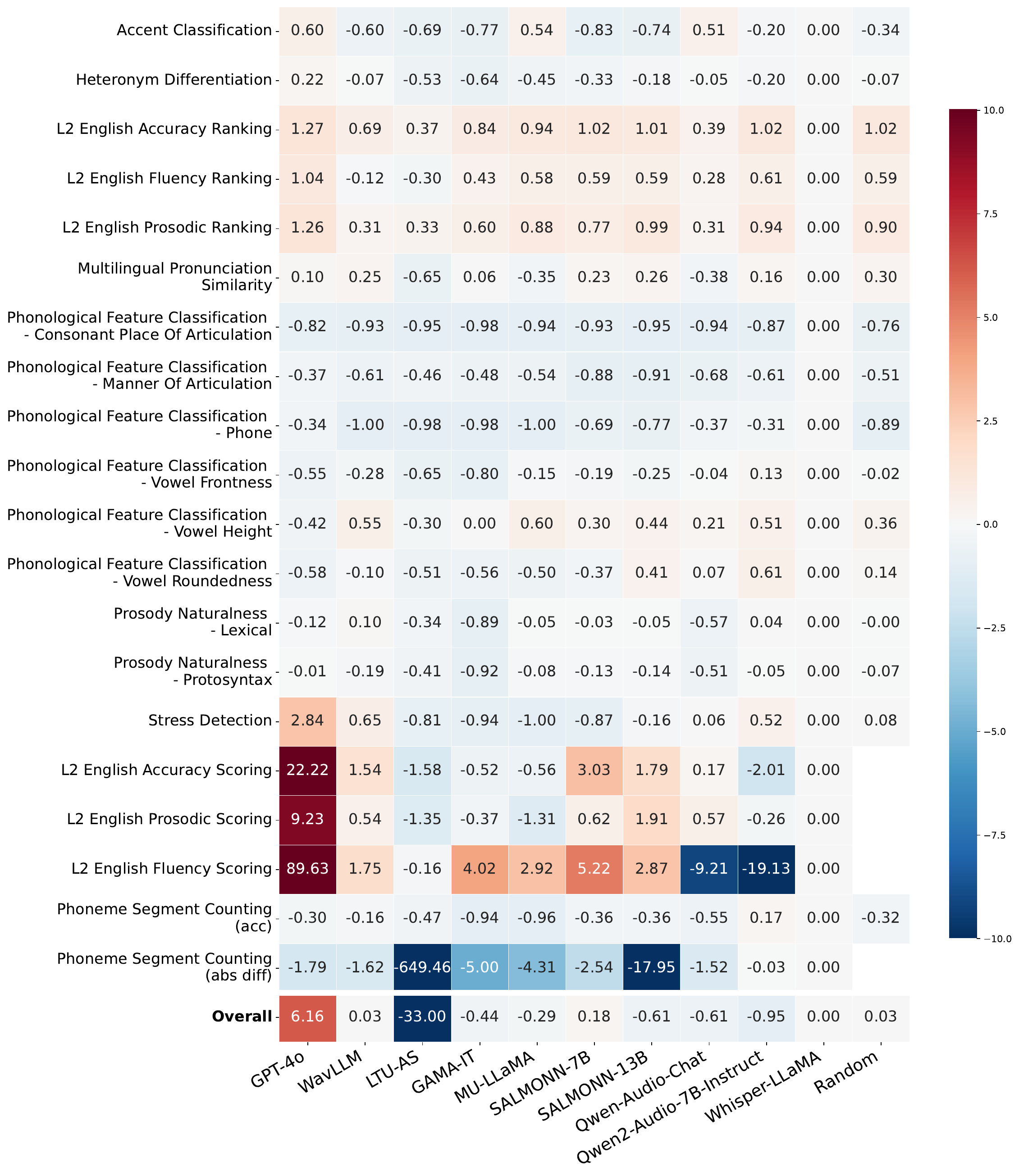} 
    \caption{Relative performance comparison of models in the~\textbf{(Speech) Phonetics, Phonology, Prosody} domain.}
    \label{fig:heatmap-speech-phonetics-phonology-prosody}
\end{figure}

\begin{figure}[htbp]
    \centering
     \includegraphics[width=\textwidth]{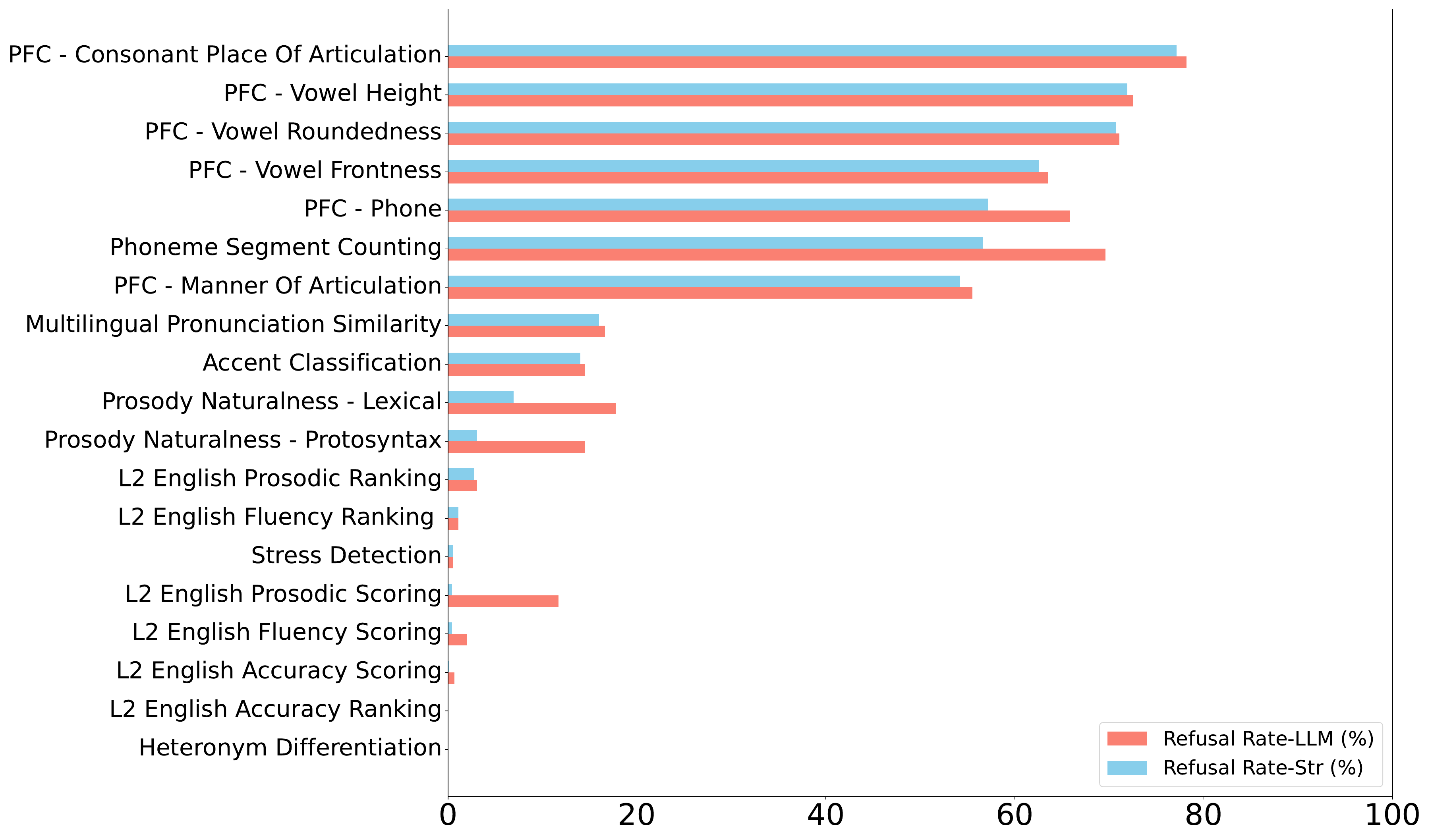} 
    \caption{Refusal rates of tasks in the~\textbf{(Speech) Phonetics, Phonology, Prosody} domain. The abbreviation ``PFC'' in the plot refers to ``Phonological Feature Classification''}
    \label{fig:rejection-speech-phonetics-phonology-prosody}
    \end{figure}

The overview, relative scores, and refusal rates of tasks about~\textbf{Phonetics, Phonology, Prosody} are demonstrated in Table~\ref{tab:tasks-speech-phonetics-phonology-prosody}, Figure~\ref{fig:heatmap-speech-phonetics-phonology-prosody} and Figure~\ref{fig:rejection-speech-phonetics-phonology-prosody} respectively. Based on our conjecture, no task in this domain raises safety or ethical concerns. Therefore, we speculate that the higher refusal rates for tasks such as ``Phonological Feature Classification'' or ``Phoneme Segment Counting'' would be attributed to the effort of post-training to avoid generating misinformation. This is supported by the fact that most audio-language models fail to achieve good results on these tasks, highlighting their difficulty for current LALMs.

Compared to the cascaded system Whisper-LLaMA, GPT-4o performs less effectively in tasks such as classification of phonological features or calculating phoneme segments, since the textual information alone provides sufficient detail for such classifications. However, GPT-4o outperforms Whisper-LLaMA and most LALMs in tasks related to accent, prosody, fluency, pronunciation, and sound stress. As these acoustic features are mainly conveyed through audio rather than text, GPT-4o demonstrates superior ability in capturing and interpreting this information. For example, in ``Stress Detection'', GPT-4o effectively identifies stress in spoken English words, achieving nearly three times the relative improvement.

Despite that GPT-4o achieves superior or comparable performance compared to baselines, the fact that Random baseline performs comparably effectively with GPT-4o indicates the failure of current LALMs on specific tasks. For ``L2 English Accuracy/Fluency/Prodosy Ranking'', GPT-4o beats this method with more than an accuracy of 50\%, as each sample in this dataset involves binary options. However, in the "Multilingual Pronunciation Similarity" task, no LALM achieves better performance than the Random baseline’s accuracy of 33\% for this triplet classification problem. The observation that such a naive random guess method exhibits comparable performance with LALMs can also be seen in a series of tasks related to phonological feature classification. The label space sizes for tasks about the consonant place of articulation, manner of articulation, phone, vowel frontness, vowel height, and vowel roundedness are 16, 12, 152, 2, 3, and 2, leading to accuracies of around 6\%, 8\%, 0.66\%, 50\%, 33\%, 50\% for the performance of Random baseline on each task. 

\begin{table}[h!]
\centering
\begin{tabular}{p{0.40\textwidth}|p{0.55\textwidth}} 
\toprule
\midrule
\textbf{Task Name} & \textbf{Task Description} \\
\midrule
- Accent
Classification & Identify the emotion of the speaker \\
\midrule 
- Heteronym Differentiation & 
Determine which sentence the pronunciation in the audio clip is more suitable for. (pairwise)\\
\midrule 
- L2 English Accuracy Ranking & \multirow{3}{0.55\textwidth}{Determine which audio is better in terms of pronunciation accuracy, speaking fluency, or prosody. (pairwise)} \\
- L2 English Fluency Ranking &  \\
- L2 English Prosodic Ranking &  \\
\midrule
- L2 English Accuracy Scoring & \multirow{3}{0.55\textwidth}{Assess the audio in terms of pronunciation accuracy, speaking fluency, or prosody. (pointwise)} \\
- L2 English Fluency Scoring &  \\
- L2 English Prosodic Scoring &  \\
\midrule 
- Multilingual Pronunciation Similarity & 
Determine which word the pronunciation of the target word is closer to, based on the three provided audio clips.\\
\midrule 
- Phonological Feature Classification & Identify the phonological features in the given audio, including place of articulation, manner of articulation, phoneme type, vowel frontness, vowel height, and vowel roundedness.\\
\midrule 
- Stress Detection & Identify the stress placement in English words. \\
\midrule 
- Prosody Naturalness & Determine which utterance one sounds more natural. (pairwise)\\
\midrule 
- Phoneme Segment Counting & Count the number of phones in the audio.\\
\midrule
\bottomrule
\end{tabular}
\vspace{10px}
\caption{Overview of tasks in the~\textbf{(Speech) Phonetics, Phonology, Prosody} domain.}
\label{tab:tasks-speech-phonetics-phonology-prosody}
\end{table}

\clearpage


\subsection{Speech Domain - Safety \& Security}

\begin{figure}[htbp]
    \centering
    \includegraphics[width=\textwidth]{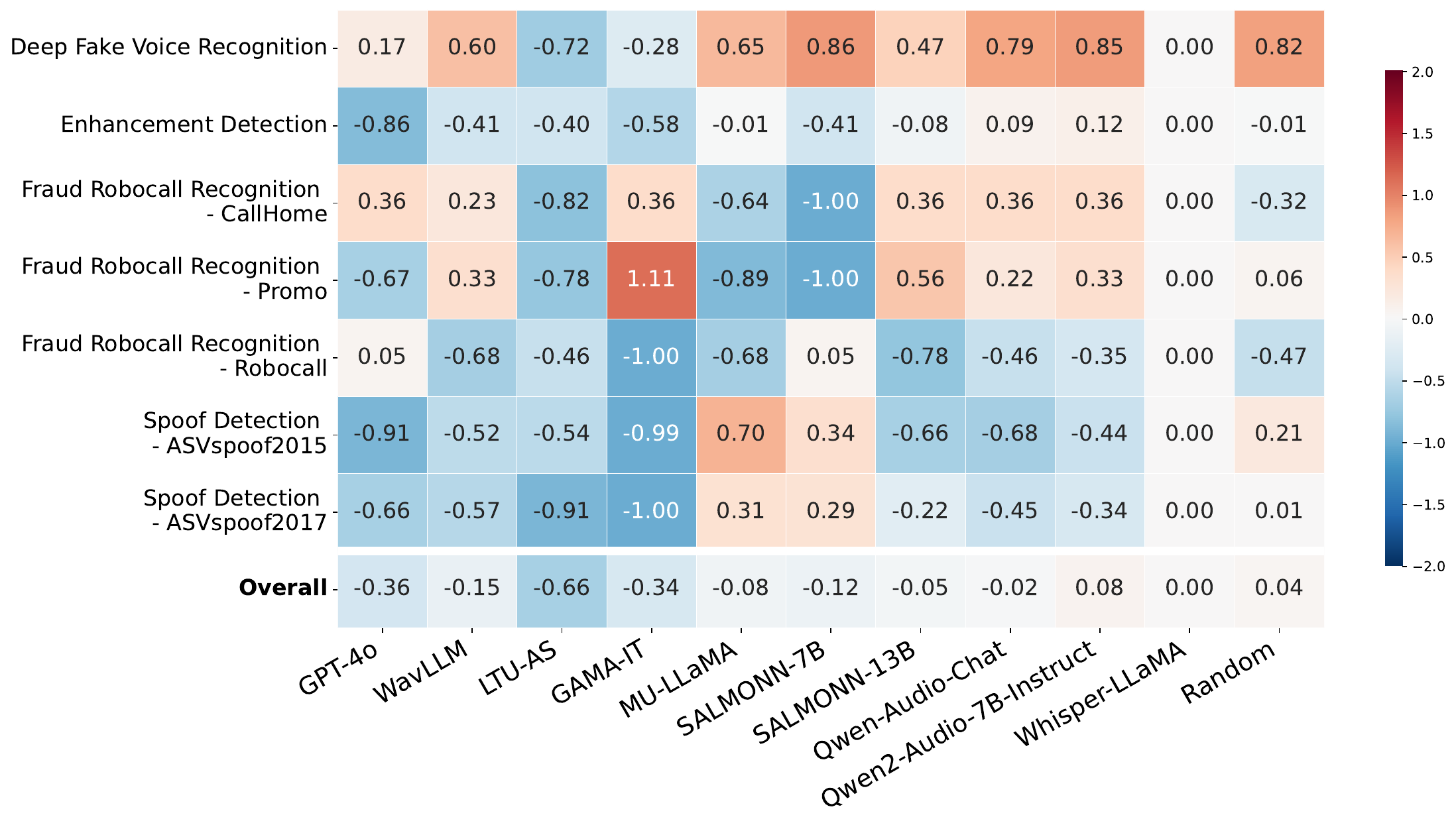} 
    \caption{Relative performance comparison of models in the~\textbf{(Speech) Safety \& Security} domain. }
    \label{fig:heatmap-speech-safety-and-security}
\end{figure}

\begin{figure}[htbp]
    \centering
     \includegraphics[width=\textwidth]{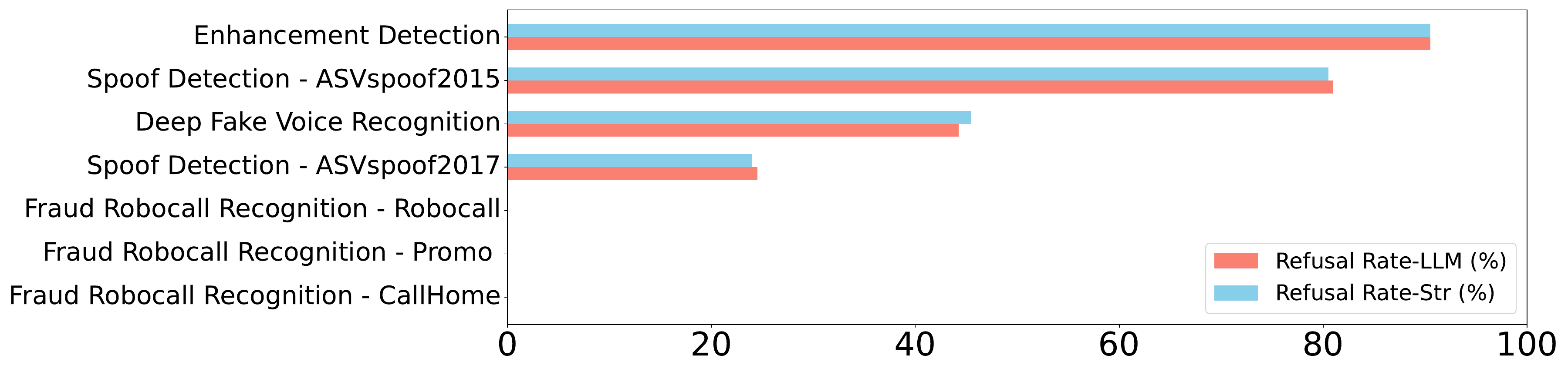} 
    \caption{Refusal rates of tasks in the~\textbf{(Speech) Safety \& Security} domain. }
    \label{fig:rejection-speech-safety-and-security}
    \end{figure}

The overview, relative scores, and refusal rates of tasks about~\textbf{Safety \& Security} are demonstrated in Table~\ref{tab:tasks-speech-safety-and-security}, Figure~\ref{fig:heatmap-speech-safety-and-security} and Figure~\ref{fig:rejection-speech-safety-and-security} respectively. Based on our conjecture, no task in this domain raises safety or ethical concerns.

For spoof detection or machine-generated audio detection, GPT-4o performs worse than most of the baseline models. Considering the absence of refusal for tasks such as ``Fraud Robocall Recognition'', we take it to look into GPT-4o performance. Interestingly, GPT-4o achieves an accuracy of 100\%, 15.79\%, and 100\% on ``CallHome'', ``Promo'', ``Robocall'' datasets, respectively, indicating its unstable performance.  In contrast, most LALMs maintain consistent performance on the ``Promo'' dataset, reflecting the comparable difficulties of these datasets. This inconsistency indicates the lack of robustness of GPT-4o when facing tasks in this domain.

\begin{table}[h!]
\centering
\begin{tabular}{p{0.35\textwidth}|p{0.60\textwidth}} 
\toprule
\midrule
\textbf{Task Name} & \textbf{Task Description} \\
\midrule
- Deep Fake Voice Recognition & \multirow{3}{0.60\textwidth}{Determine if the speech is generated or altered by machine}  \\
- Enhancement Detection & \\
- Spoof Detection & \\
\midrule
- Fraud Robocall Recognition & Determine if the phone call is a spam.  \\

\midrule 
\bottomrule
\end{tabular}
\vspace{10px}
\caption{Overview of tasks in the~\textbf{(Speech) Safety \& Security} domain.}
\label{tab:tasks-speech-safety-and-security}
\end{table}

\vspace{-20px}

\clearpage

\subsection{Speech Domain - Speaker \& Language}

\begin{figure}[htbp]
    \centering
    \includegraphics[width=\textwidth]{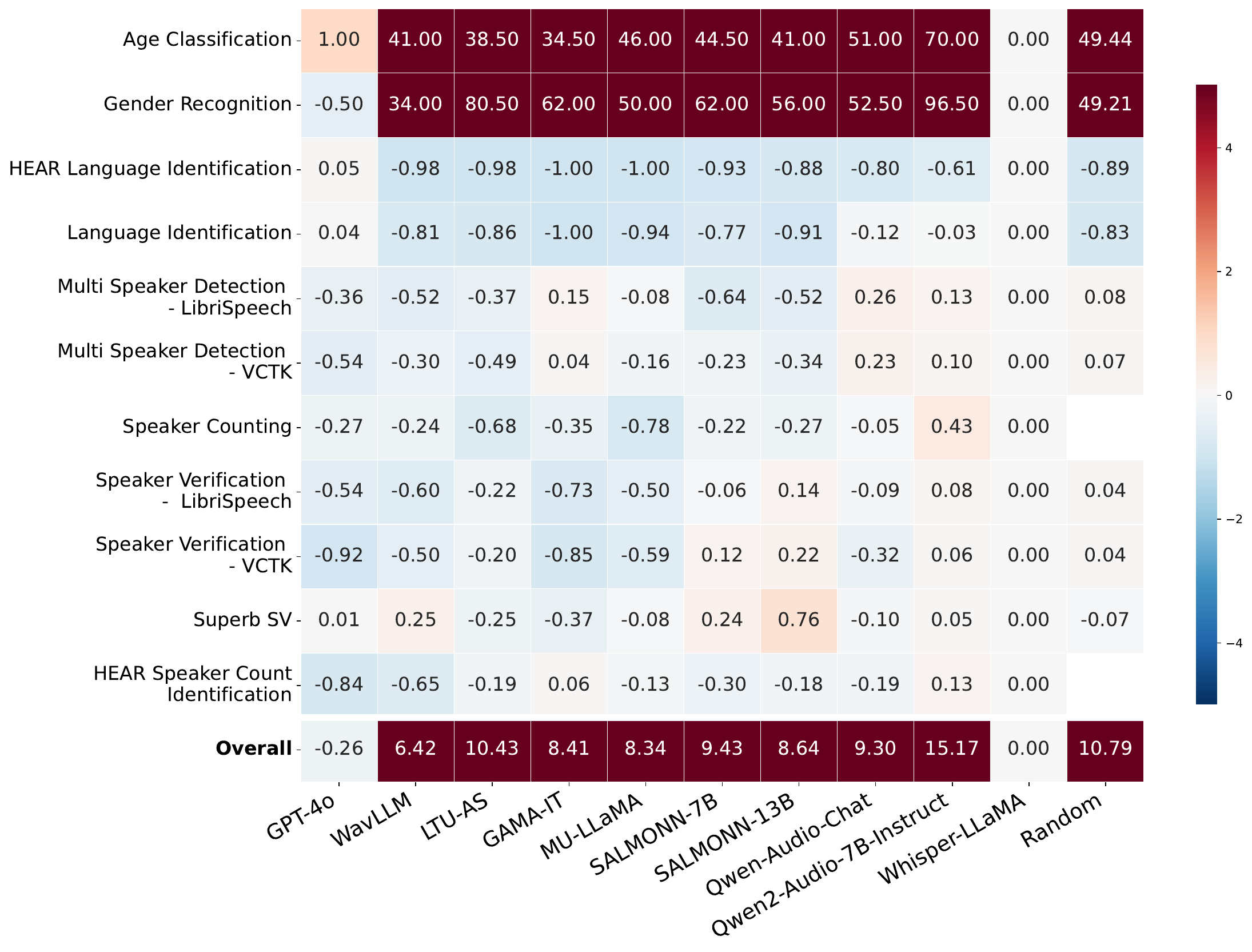} 
    \caption{Relative performance comparison of models in the~\textbf{(Speech) Speaker \& Language} domain. }
    \label{fig:heatmap-speech-speaker-and-language}
\end{figure}

\begin{figure}[htbp]
    \centering
     \includegraphics[width=\textwidth]{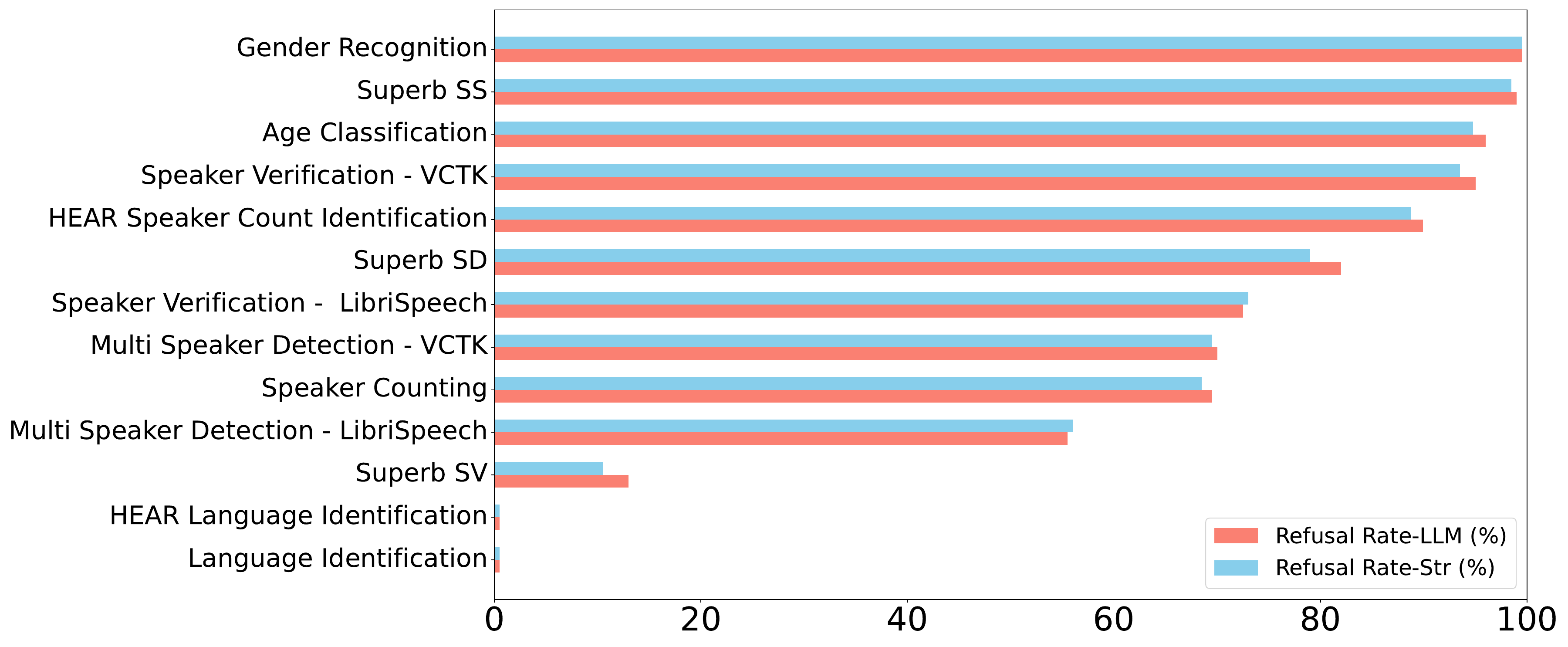} 
    \caption{Refusal rates of tasks in the~\textbf{(Speech) Speaker \& Language} domain. }
    \label{fig:rejection-speech-speaker-and-language}
    \end{figure}

The overview, relative scores, and refusal rates of tasks about~\textbf{Speaker \& Language} are demonstrated in Table~\ref{tab:tasks-speech-speaker-and-language}, Figure~\ref{fig:heatmap-speech-speaker-and-language} and Figure~\ref{fig:rejection-speech-speaker-and-language} respectively. Based on our conjecture, tasks in this domain often raise some safety concerns. Tasks such as Age or Gender Classification are a type of Ungrounded Inference, with nearly complete refusals for all testing samples. Also, Speaker Verification is fundamentally to do speaker identification. Furthermore, GPT-4o shows a significantly high refusal rate in ``SUPERB SS''. We attribute this refusal to potential concerns about the generation of copyrighted materials, as GPT-4o avoids reproducing speech beyond the preset voices. Interestingly, although ``Speaker Verification'' and ``SUPERB SV'' are fundamentally the same, GPT-4o exhibits distinct behavior toward these tasks, tending to refuse the former while responding to the latter. We speculate that this discrepancy is due to the differing instructions used in these tasks, suggesting the sensitivity of GPT-4o's refusal mechanism.

In sum, since the tasks about~\textbf{Speaker \& Language} are usually associated with high safety concerns, making GPT-4o refuses to comply in  most cases. The exception tasks are tasks about Language Identification and ``superbSV''. Because there is a notable gap between languages involved in ``HEAR Language Identification'' and ``Language Identification'', the cascaded model can solve it by pure text transcription without relying on accent or pronunciation, making Whisper-LLaMA a strong baseline. Although Whisper-LLaMA outperforms all LALMs, GPT-4 still achieves better performance, reflecting its notable capabilities for identifying language.

\begin{table}[h!]
\centering
\begin{tabular}{p{0.4\textwidth}|p{0.55\textwidth}} 
\toprule
\midrule
\textbf{Task Name} & \textbf{Task Description} \\
\midrule
- Age classification & Identify the age of the speaker.\\
\midrule
- Gender classification & Identify the gender of the speaker. \\
\midrule
- HEAR Language Identification & \multirow{2}{0.60\textwidth}{Identify the language spoken in the audio.}  \\
- Language Identification & \\
\midrule
- Multi Speaker Detection & Determine if multiple speakers are present. \\
\midrule
- Speaker Counting &  \multirow{2}{0.55\textwidth}{Count the number of speakers.} \\
- HEAR Speaker Count Identification & \\
\midrule
- Speaker Verification & \multirow{2}{0.55\textwidth}{Determine if the given speech clips are spoken by the same speakers} \\
- SUPERB SV & \\
\midrule
- Target Speaker ASR & \multirow{2}{0.55\textwidth}{Write down the transcription of each speaker of a clip multi-speaker speech.} \\
- SUPERB SD & \\
\midrule 
\bottomrule
\end{tabular}
\vspace{10px}
\caption{Overview of tasks in the~\textbf{(Speech) Speaker \& Language} domain.}
\label{tab:tasks-speech-speaker-and-language}
\end{table}

\clearpage


\subsection{Speech Domain - Speech Enhancement}

\begin{figure}[htbp]
    \centering
     \includegraphics[width=\textwidth]{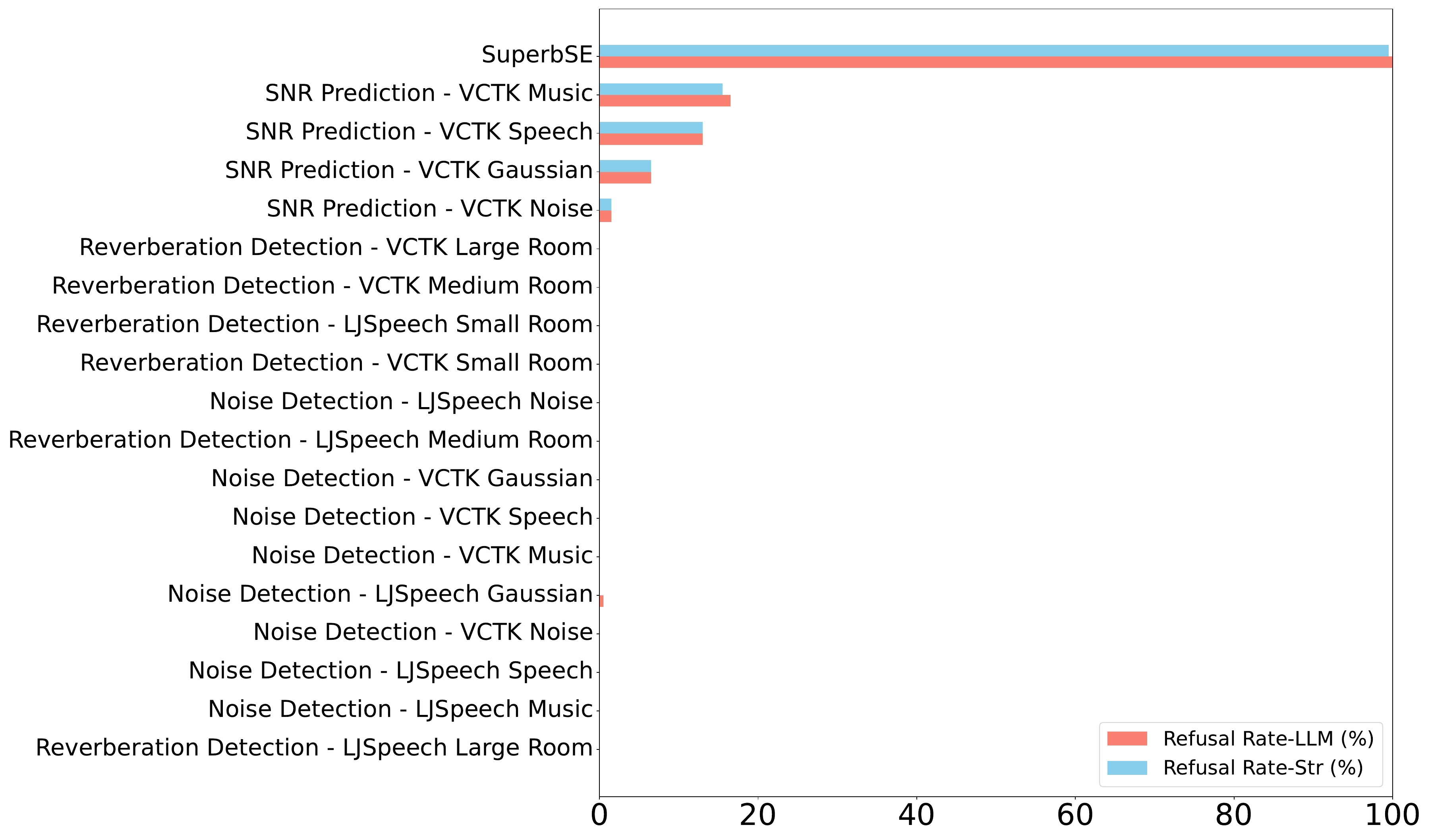} 
    \caption{Refusal rates of tasks in the~\textbf{(Speech) Speech Enhancement} domain. }
    \label{fig:rejection-speech-speech-enhancement}
    \end{figure}

\begin{figure}[htbp]
    \centering
    \includegraphics[width=\textwidth]{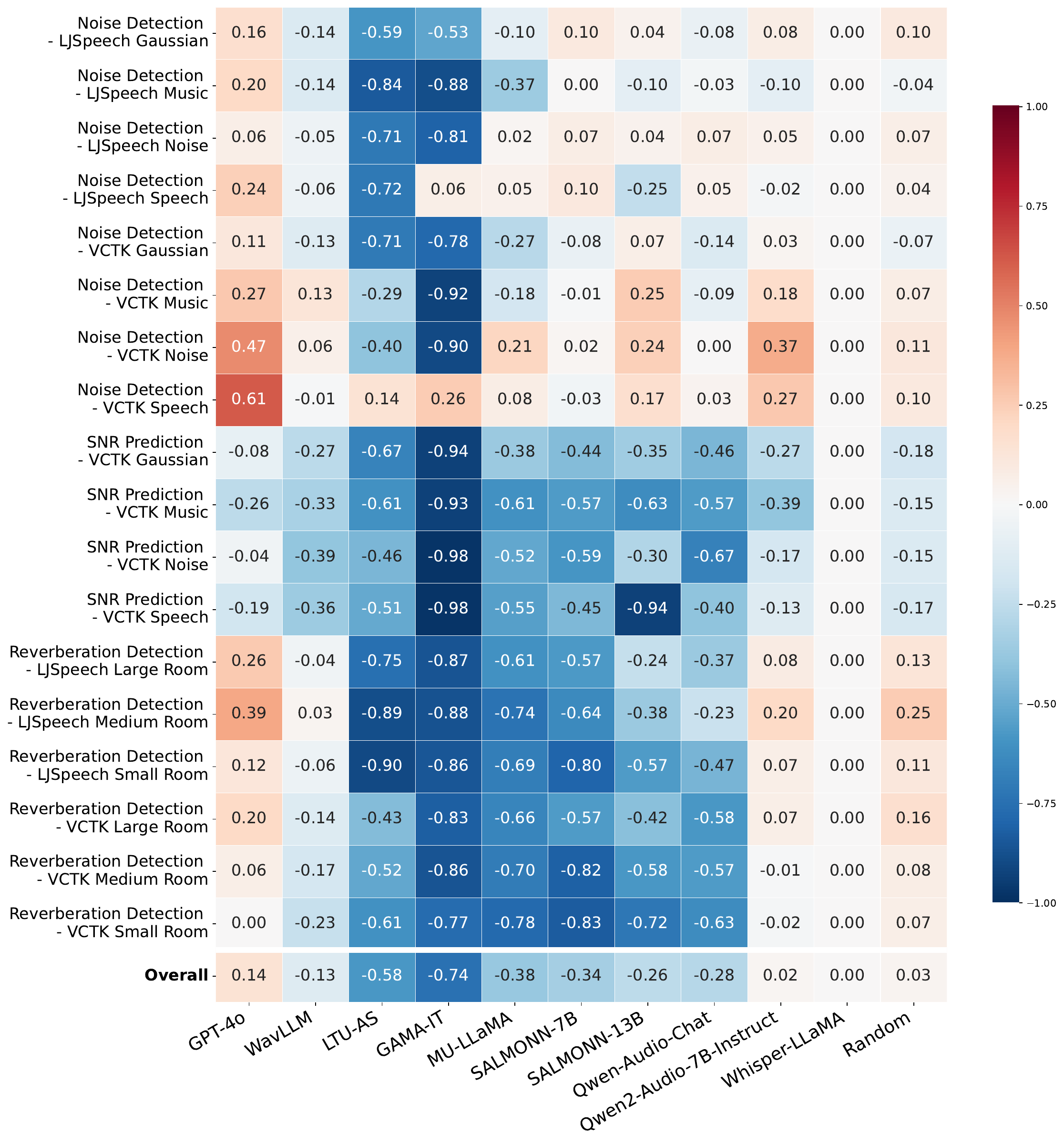} 
    \caption{Relative performance comparison of models in the~\textbf{(Speech) Speech Enhancement} domain. }
    \label{fig:heatmap-speech-speech-enhancement}
\end{figure}

The overview, relative scores, and refusal rates of tasks about~\textbf{Speech Enhancement} are demonstrated in Table~\ref{tab:tasks-speech-speech-enhancement}, Figure~\ref{fig:heatmap-speech-speech-enhancement} and Figure~\ref{fig:rejection-speech-speech-enhancement} respectively. Based on our conjecture, the primary safety concern with ``SUPERB SE'' appears to be the generation of copyrighted materials, as it involves reproducing input audio.

In tasks such as detecting noise and reverberation, GPT-4o generally outperforms baseline audio-language models and Whipser-LLaMA. However, GPT-4o performs less effectively than Whisper-LLaMA specifically in SNR prediction. We attribute this to the significant gap between the options in these tasks, such as zero, five, or ten, enabling textual LLMs to infer the correct answer based on the fluency and accuracy of ASR transcriptions without hearing audio.

\begin{table}[h!]
\centering
\begin{tabular}{p{0.35\textwidth}|p{0.60\textwidth}} 
\toprule
\midrule
\textbf{Task Name} & \textbf{Task Description} \\
\midrule
- SUPERB SE & Enhance the speech.\\
\midrule
- Noise Detection & Determine if the audio is clean or noisy.\\
\midrule 
- SNR Prediction & Predict the Signal-to-Noise Ratio (SNR) for the given speech.\\
\midrule 
- Reverberation Detection & 
Determine if the speech contains reverberation noise in the provided environment. \\
\midrule 
\bottomrule
\end{tabular}
\vspace{10px}
\caption{Overview of tasks in the~\textbf{(Speech) Speech Enhancement} domain.}
\label{tab:tasks-speech-speech-enhancement}
\end{table}

\clearpage


\subsection{Speech Domain - Speech Recognition}

\begin{figure}[htbp]
    \centering
    \includegraphics[width=\textwidth]{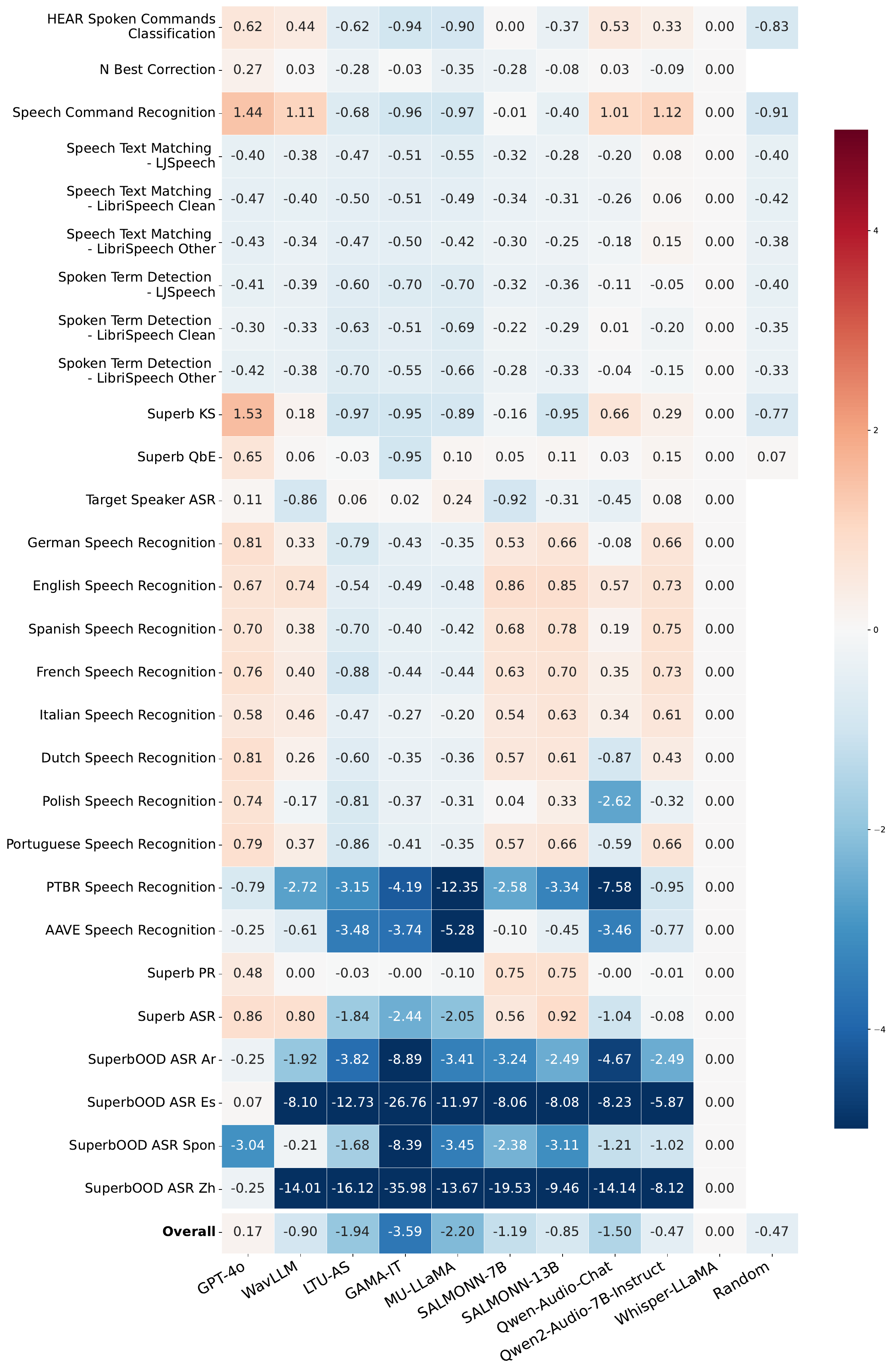} 
    \caption{Relative performance comparison of models in the~\textbf{(Speech) Speech Recognition} domain. }
    \label{fig:heatmap-speech-speech-recognition}
\end{figure}

\begin{figure}[htbp]
    \centering
     \includegraphics[width=\textwidth]{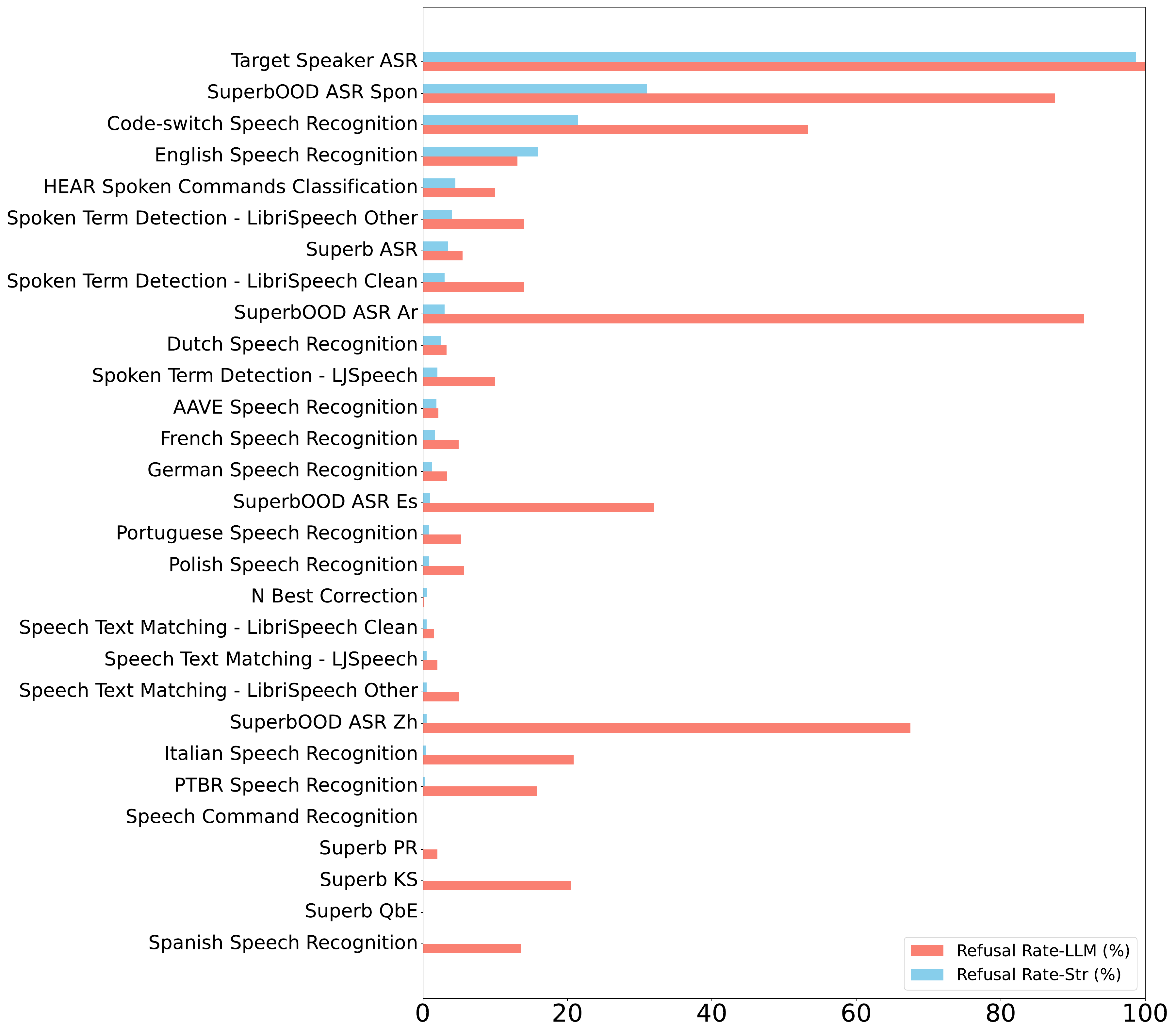} 
    \caption{Refusal rates of tasks in the~\textbf{(Speech) Speech Recognition} domain. }
    \label{fig:rejection-speech-speech-recognition}
    \end{figure}

The overview, relative scores, and refusal rates of tasks about~\textbf{Speech Recognition} are demonstrated in Table~\ref{tab:tasks-speech-speech-recognition}, Figure~\ref{fig:heatmap-speech-speech-recognition} and Figure~\ref{fig:rejection-speech-speech-recognition} respectively. Based on our conjecture, there is no task associated with safety concerns. After manual inspection, we find that the significant gap between refusal rates determined by string match methods and LLaMA-3.1-8B-Instruct is attributed to the misjudgment  of the LLM. Hence, the refusal rate in Figure~\ref{fig:rejection-speech-speech-recognition} determined by string matching methods is more accurate and reliable.  

Intuitively, Whisper-LLaMA serves as a strong baseline for speech recognition, with Whisper focusing on transcribing speech and LLaMA acting as a transcription corrector. However, GPT-4o surpasses Whisper-LLaMA in recognizing German, English, Spanish, French, Italian, Dutch, Polish, and Portuguese speech, highlighting GPT-4o's impressive multilingual speech recognition capabilities. Notably, while most audio-language models perform worse than Whisper-LLaMA in Polish speech recognition, GPT-4o achieves a significantly lower word error rate.

When it comes to other tasks related to ASR, GPT-4o beats the baselines in tasks like keyword spotting, spoken commands classification, ASR transcription correction, and query by example. Conversely, on tasks like ``Multi Speaker Detection'', ``Speech Text Matching'', GPT-4o performs less effectively than other audio-language models, with Whisper-LLaMA achieving better results compared to most of the audio-langauge models. Surprisingly, despite there being no audio input for the N-best ASR hypothesis correction task, GPT-4o still outperforms Whisper-LLaMA, reflecting the the wealth of textual knowledge owned by this multi-modal model.

\begin{table}[h!]
\centering
\begin{tabular}{p{0.45\textwidth}|p{0.50\textwidth}} 
\toprule
\midrule
\textbf{Task Name} & \textbf{Task Description} \\
\midrule
- HEAR Spoken Commands Classification & \multirow{2}{0.50\textwidth}{Identify the keyword delivered in speech.} \\
- Speech Command Recognition & \\
- SUPERB KS & \\
\midrule
- N Best Correction & Correct ASR transcription with n-best hypotheses.\\
\midrule
- Speech Text Matching & Determine if speech and text correspond to each other.\\
\midrule
- SUPERB QbE & Determine if the word spoken in the brief utterance is present in the longer one.\\
\midrule 
- Target Speaker ASR & Transcribe the speech in a multi-speaker audio delivered by the speaker in another single-speaker audio.\\
\midrule
- Speech Recognition & \multirow{3}{0.50\textwidth}{Transcribe speech into text.}\\
- SUPERB ASR & \\
- SUPERB OOD ASR & \\
\midrule
- SUPERB PR & Transcribes speech into phoneme. \\
\midrule 
\bottomrule
\end{tabular}
\vspace{10px}
\caption{Overview of tasks in the~\textbf{(Speech) Speech Recognition} domain.}
\label{tab:tasks-speech-speech-recognition}
\end{table}

\clearpage


\subsection{Speech Domain - Speech, Voice, Hearing Disorder}

\begin{figure}[htbp]
    \centering
    \includegraphics[width=\textwidth]{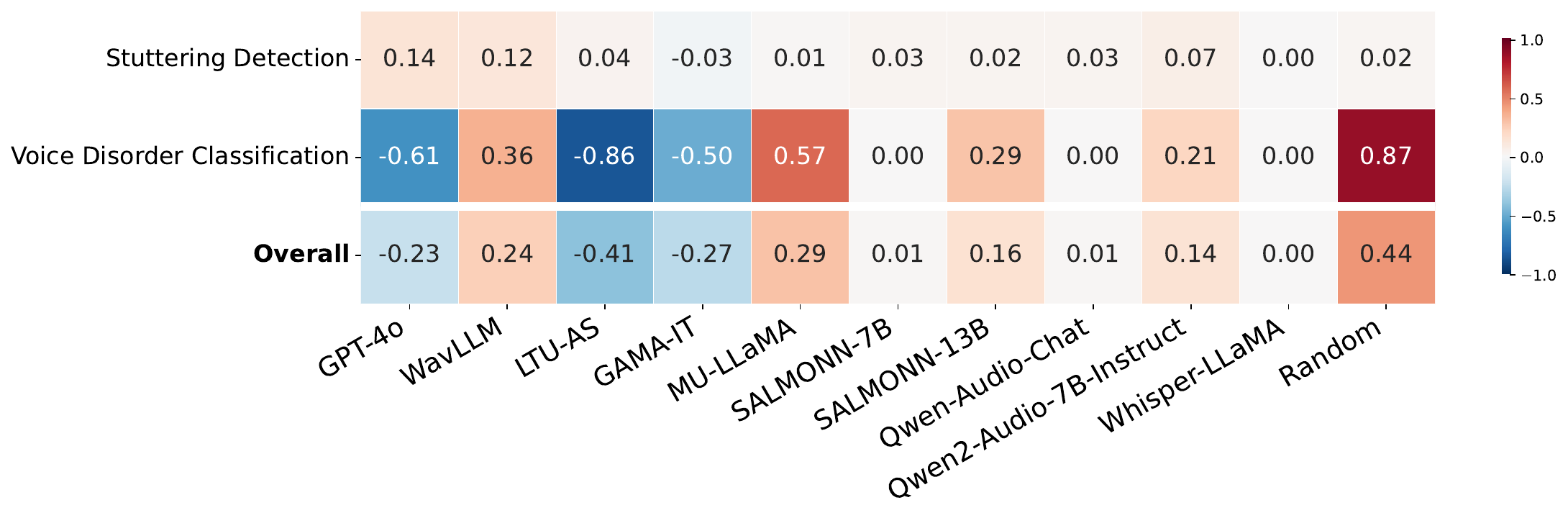} 
    \caption{Relative performance comparison of models in the~\textbf{(Speech) Speech, Voice, Hearing Disorder} domain. }
    \label{fig:heatmap-speech-speech-voice-hearing-disorder}
\end{figure}

\begin{figure}[htbp]
    \centering
     \includegraphics[width=\textwidth]{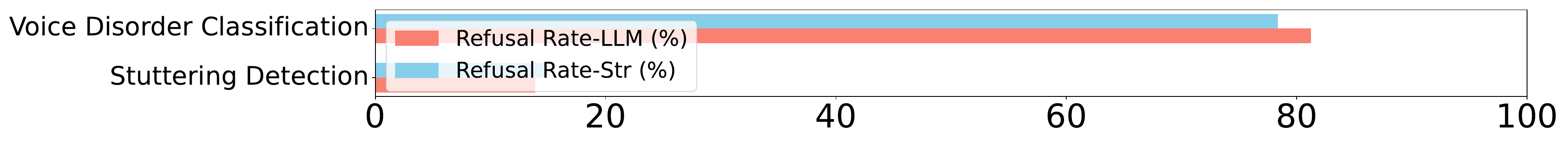} 
    \caption{Refusal rates of tasks in the~\textbf{(Speech)  Speech, Voice, Hearing Disorder} domain. }
    \label{fig:rejection-speech-speech-voice-hearing-disorder}
    \end{figure}

The overview, relative scores, and refusal rates of tasks about~\textbf{Speech, Voice, Hearing Disorder} are demonstrated in Table~\ref{tab:tasks-speech-speech-voice-hearing-disorder}, Figure~\ref{fig:heatmap-speech-speech-voice-hearing-disorder} and Figure~\ref{fig:rejection-speech-speech-voice-hearing-disorder} respectively. 

GPT-4o outperforms LALM baselines and Whisper-LLaMA In ``Stuttering Detection'', showing its superior ability to handle speech characteristics and effectively identify speech disfluencies. However, for ``Voice Disorder Classification'', even MU-LLaMA, which achieves the best performance in this task, only reaches an accuracy of 21.5\%, falling below the accuracy of 24.5\% from Random baseline. This indicates that current LALMs struggle to accurately identify voice disorders.
\begin{table}[h!]
\centering
\begin{tabular}{p{0.35\textwidth}|p{0.6\textwidth}} 
\toprule
\midrule
\textbf{Task Name} & \textbf{Task Description} \\
\midrule
- Voice Disorder Classification & Diagnose the voice to determine whether the speaker is affected by hyperkinetic, hypokinetic, or flux laryngitis. \\
\midrule
- Stuttering Detection & Determine if any stuttering in the speech.\\
\midrule 
\bottomrule
\end{tabular}
\vspace{10px}
\caption{Overview of tasks in the~\textbf{(Speech) Speech, Voice, Hearing Disorder} domain.}
\label{tab:tasks-speech-speech-voice-hearing-disorder}
\end{table}

\clearpage


\subsection{Speech Domain - Spoken Language Understanding}

\begin{figure}[htbp]
    \centering
    \includegraphics[width=\textwidth]{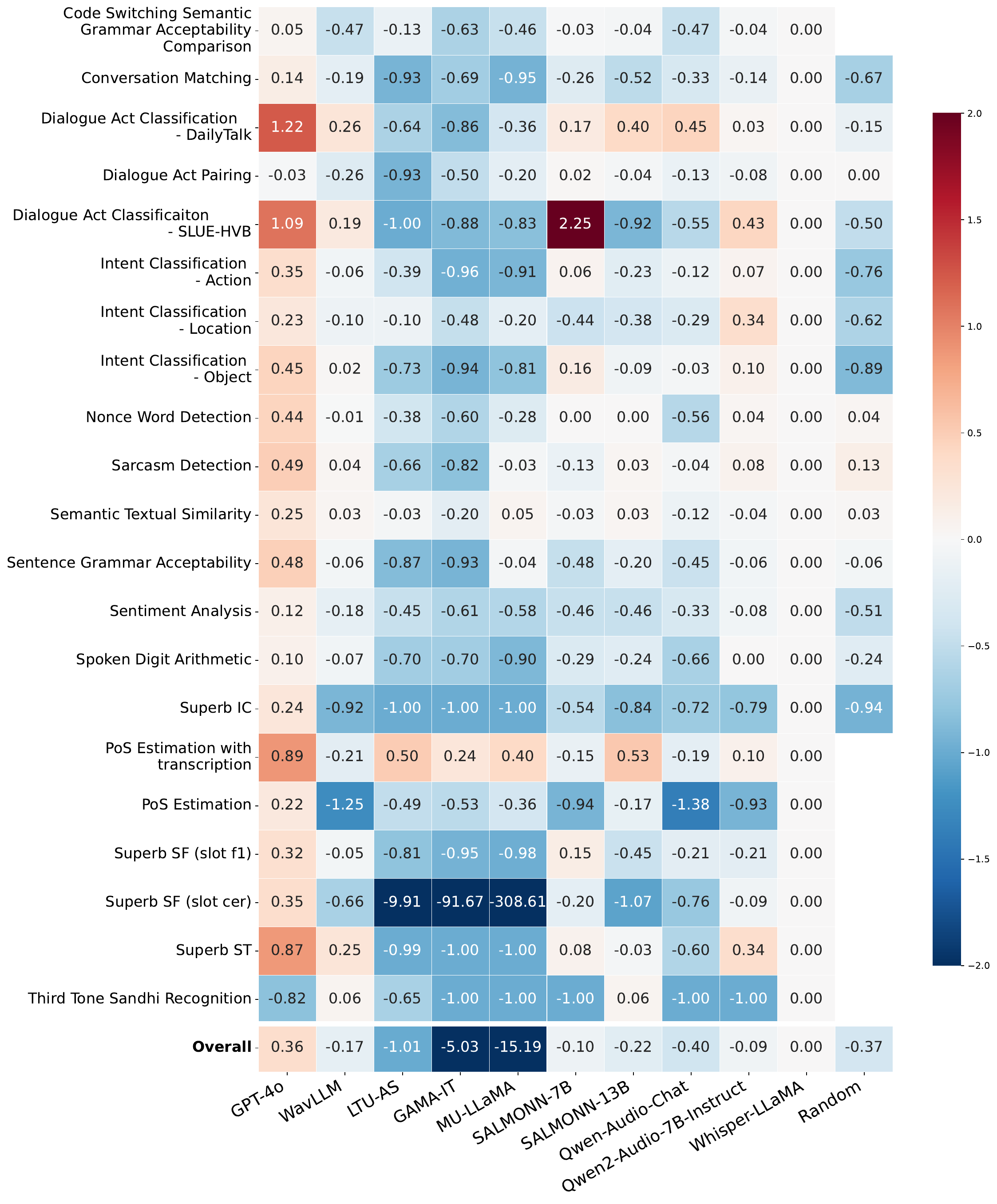} 
    \caption{Relative performance comparison of models in the~\textbf{(Speech) Spoken Language Understanding} domain. }
    \label{fig:heatmap-speech-spoken-language-understanding}
\end{figure}

\begin{figure}[htbp]
    \centering
     \includegraphics[width=\textwidth]{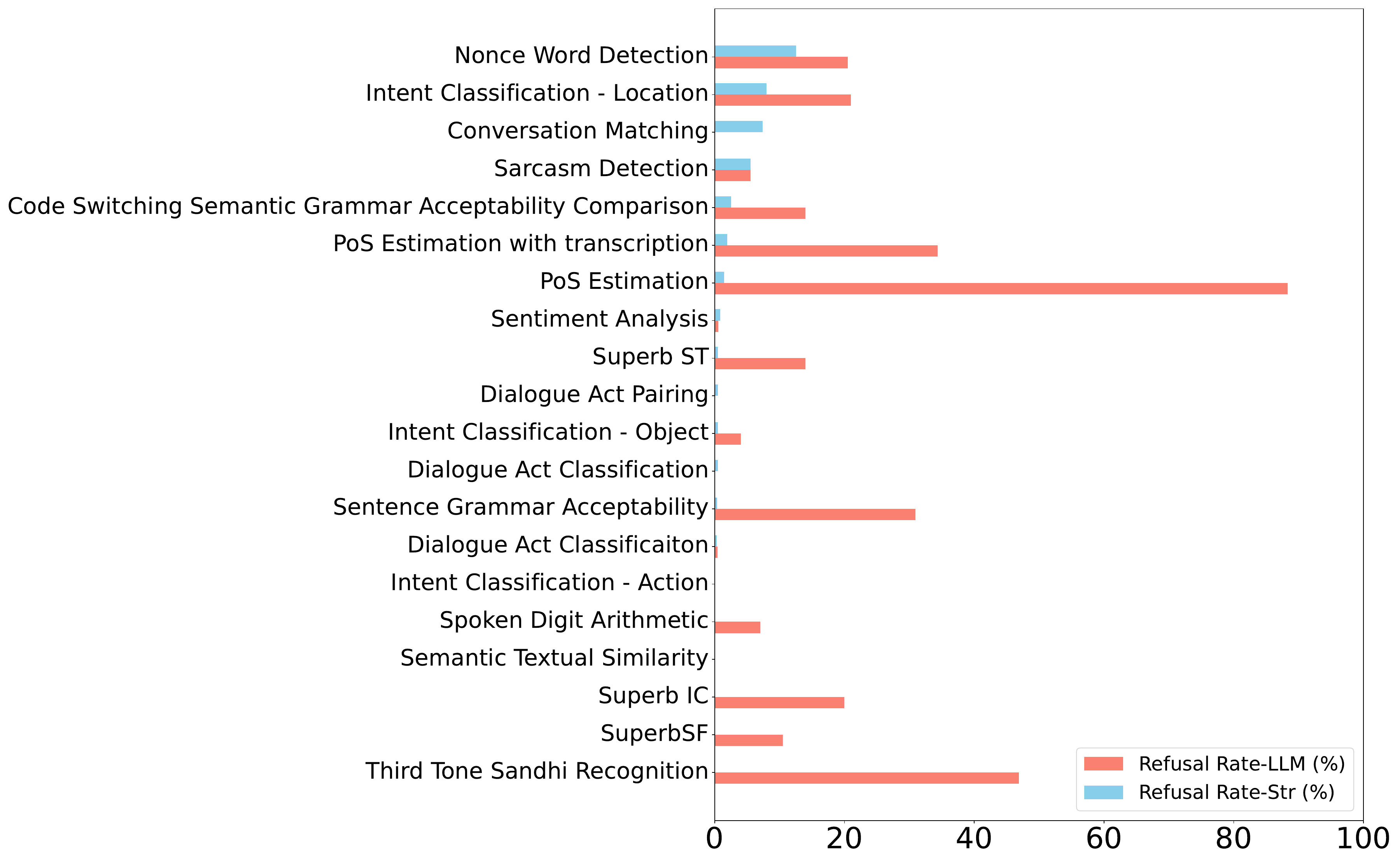} 
    \caption{Refusal rates of tasks in the~\textbf{(Speech) Spoken Language Understanding} domain. }
    \label{fig:rejection-speech-spoken-language-understanding}
    \end{figure}

The overview, relative scores, and refusal rates of tasks about~\textbf{Spoken Language Understanding} are demonstrated in Table~\ref{tab:tasks-speech-spoken-language-understanding}, Figure~\ref{fig:heatmap-speech-spoken-language-understanding} and Figure~\ref{fig:rejection-speech-spoken-language-understanding} respectively. Based on our conjecture, no tasks in this domain are associated with safety concerns. After manual inspection, we find that the significant gap between refusal rates determined by string match methods and LLaMA-3.1-8B-Instruct is attributed to the misjudgment of the LLM. Hence, the refusal rate in Figure~\ref{fig:rejection-speech-spoken-language-understanding} determined by string matching methods is more accurate and reliable. 

Roughly speaking, GPT-4o outperforms all LALM baselines and Whipser-LLaMA on tasks in this domain. Notably, Whisper-LLaMA beats most LALMs but not GPT-4o. This result reflects GPT-4o's ability to effectively comprehend and process complex spoken language inputs, including subtle linguistic and acoustics nuances and variations in speech. It also highlights its robust generalization skills across diverse speech-related tasks.

\begin{table}[h!]
\centering
\begin{tabular}{p{0.4\textwidth}|p{0.55\textwidth}} 
\toprule
\midrule
\textbf{Task Name} & \textbf{Task Description} \\
\midrule
- Code Switching Semantic Grammar Acceptability Comparison & 
Judge if the selected utterance is clearer, more fluent, or more grammatical than the other. \\
\midrule
- Conversation Matching & Choose a response based on the provided audio.\\
\midrule
- Dialogue Act Classification & 
Identify the speaker's communicative dialogue acts. \\
\midrule
- Dialogue Act Pairing & 
Determine if the two dialogue acts are identical.\\
\midrule 
- Intent Classification &  \multirow{2}{0.50\textwidth}{Identify the purpose of the speech.}\\
- SUPERB IC & \\
\midrule 
- Nonce Word Detection & Determine which spoken word is fake. (pairwise) \\
\midrule 
- Sarcasm Detection & 
Determine if sarcasm or irony is employed in the speech.\\
\midrule 
- Semantic Textual Similarity & 
Given two pairs of utterances, identify which pair is more similar.\\
\midrule 
- Sentence Grammar Acceptability & 
Determine which utterance is grammatically accurate. (pairwise)\\
\midrule 
- Sentiment Analysis &  Identify the sentiment of the speech.\\
\midrule 
- Spoken Digit Arithmetic &  Identify the result of spoken digit arithmetic.\\
\midrule 
- PoS Estimation with transcription  & Transcribe the audio and then identify the corresponding Part-of-Speech (POS) tags of the words in the utterance. \\
\midrule 
- PoS Estimation &  Identify the corresponding Part-of-Speech (POS) tags of the words in the utterance. \\
\midrule 
- SUPERB SF & Identify the slot values within the spoken statements, by using the provided intent and related slot types, \\
\midrule 
- SUPERB ST & Translate the speech phrase into the specific language. \\
\midrule 
- Third Tone Sandhi Recognition & 
Write down the position(s) of the character(s) with a third tone sandhi.\\

\bottomrule
\end{tabular}
\vspace{10px}
\caption{Overview of tasks in the~\textbf{(Speech) Spoken Language Understanding} domain.}
\label{tab:tasks-speech-spoken-language-understanding}
\end{table}

\clearpage

\subsection{Music Domain - Harmony \& Pitch}

\begin{figure}[htbp]
    \centering
    \includegraphics[width=\textwidth]{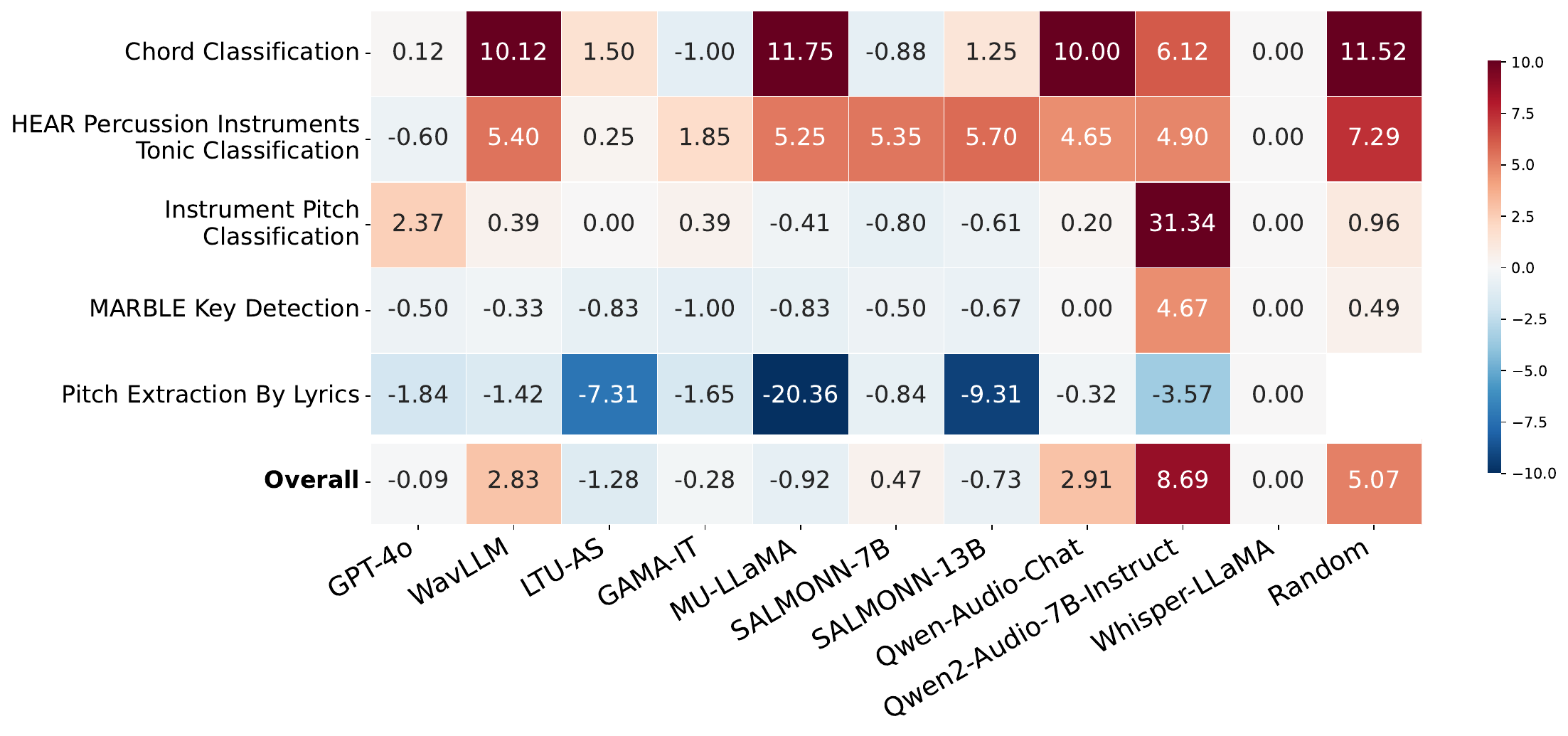} 
    \caption{Relative performance comparison of models in the~\textbf{(Music) Harmony \& Pitch} domain. }
    \label{fig:heatmap-audio-harmony-pitch}
\end{figure}

\begin{figure}[htbp]
    \centering
     \includegraphics[width=\textwidth]{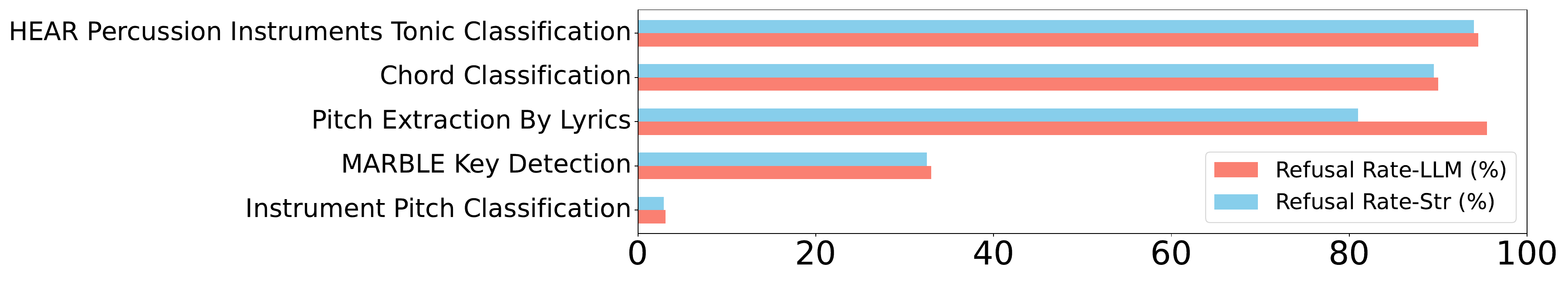} 
    \caption{Refusal rates of tasks in the~\textbf{(Music) Harmony \& Pitch} domain. }
    \label{fig:rejection-audio-harmony-pitch}
    \end{figure}

The overview, relative scores, and refusal rates of tasks about~\textbf{Harmony \& Pitch} are demonstrated in Table~\ref{tab:tasks-audio-harmony-pitch}, Figure~\ref{fig:heatmap-audio-harmony-pitch} and Figure~\ref{fig:rejection-audio-harmony-pitch} respectively. GPT-4o shows a significantly high refusal rate in tasks like ``HEAR Percussion Instruments Tonic Classification'', ``Chord Classification'', and ``Pitch Extraction By Lyrics''. These tasks are to identify the tonic or chord of music and appear to be not associated with the safety concerns outlined in the official report. Besides, GPT-4o tends to fulfill the requests in ``Instrument Pitch Classification'' but refuses to respond ``Pitch Extraction By Lyrics''. These two tasks are fundamentally the same, but the latter is more challenging due to the absence of predefined options and the potential for a sequence of pitches. These observations suggest that the reason GPT-4o refuse to respond is that the tasks are too hard for GPT-4o to complete accurately so it refused to respond to avoid generating misinformation.

Due to the high refusal rate of GPT-4o, WavLLM, Qwen-Audio-Chat, and Qwen2-Audio-7B-Instruct outperform it in the majority of tasks. Surprisingly, although the transcription produced by Whisper mainly focuses on language instead of acoustics, Whisper-LLaMA outperforms LALMs in ``Pitch Extraction By Lyrics''. We speculate that the knowledge stored in LLaMA acquired during the pretraining stage includes the textual content of music samples in this dataset. This enables Whisper-LLaMA to accurately generate pitch sequences according to the transcribed lyrics, even without relying on acoustic features. Besides, Random baseline achieve superior or comparable performance than LALMs, indicating the room for improvement of LALMs.

\begin{table}[h!]
\centering
\begin{tabular}{p{0.35\textwidth}|p{0.6\textwidth}} 
\toprule
\midrule
\textbf{Task Name} & \textbf{Task Description} \\
\midrule
- Chord Classification & Determine if a tune is major or minor. \\
\midrule
- HEAR Percussion Instruments Tonic Classification & Identify the tonic note. \\
\midrule
- Instrument Pitch Classification &  Identify the pitch in the music.\\
\midrule
- MARBLE Key Detection & Identify the musical key of the given audio.\\
\midrule
- Pitch Extraction By Lyrics &  Write down a sequence of pitches in the music.\\
\midrule 
\bottomrule
\end{tabular}
\vspace{10px}
\caption{Overview of tasks in the~\textbf{(Music) Harmony \& Pitch} domain.}
\label{tab:tasks-audio-harmony-pitch}
\end{table}

\clearpage
\vspace{10px}
\clearpage


\subsection{Music Domain - Music Classification}

\begin{figure}[htbp]
    \centering
    \includegraphics[width=\textwidth]{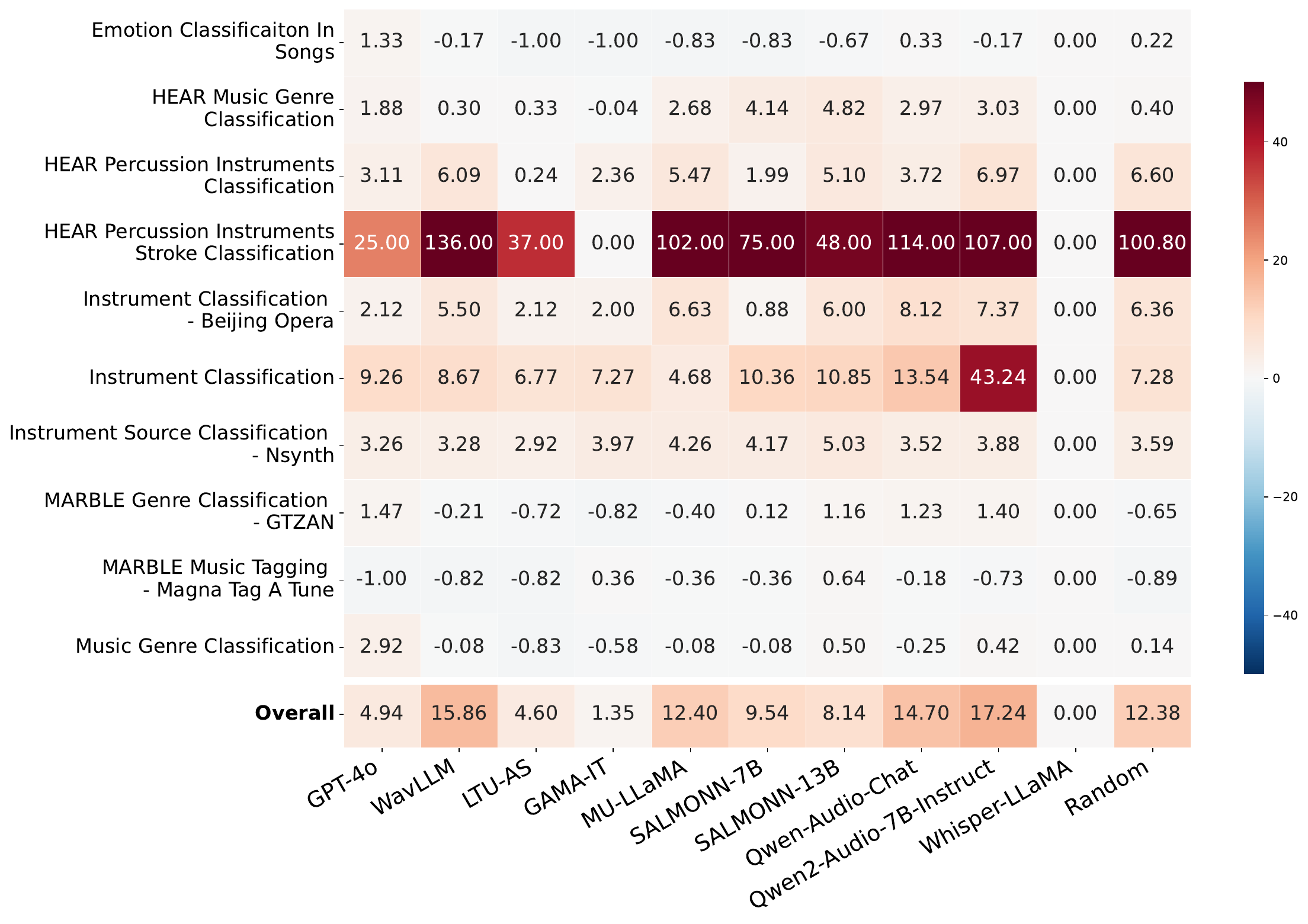} 
    \caption{Relative performance comparison of models in the~\textbf{(Music) Music Classification} domain. }
    \label{fig:heatmap-audio-music-classification}
\end{figure}

\begin{figure}[htbp]
    \centering
     \includegraphics[width=\textwidth]{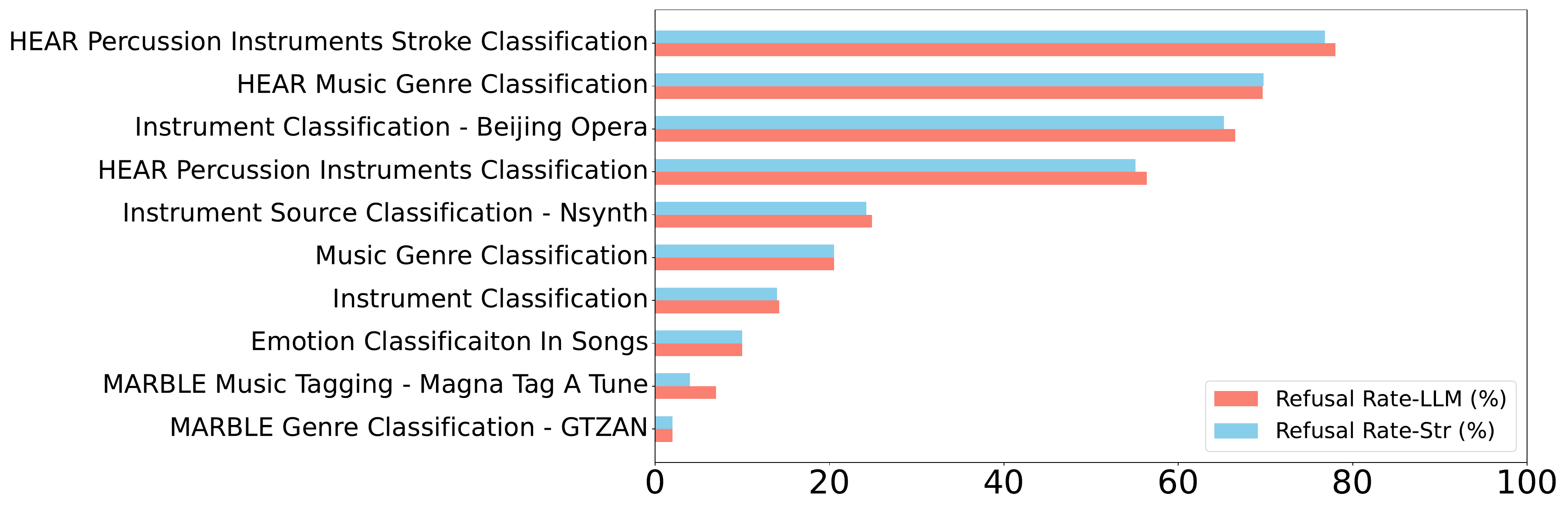} 
    \caption{Refusal rates of tasks in the~\textbf{(Music) Music Classification} domain. }
    \label{fig:rejection-audio-music-classification}
    \end{figure}

The overview, relative scores, and refusal rates of tasks about~\textbf{Music Classification} are demonstrated in Table~\ref{tab:tasks-audio-music-classification}, Figure~\ref{fig:heatmap-audio-music-classification} and Figure~\ref{fig:rejection-audio-music-classification} respectively. GPT-4o shows a significantly high refusal rate in tasks like ``HEAR Percussion Instruments Stroke Classification'', ``HEAR Music Genre Classification'', and so on. As these tasks are to identify the genres of music and appear to be not associated with the safety concerns outlined in the official report, we speculate that the reason GPT-4o refuses to respond is that the tasks are too hard for GPT-4o to complete accurately so that it refused to respond to avoid generating misinformation.

As the main purpose of Whisper is to transcribe speech, the acoustic features are lost during transcription, making Whisper-LLaMA an ineffective baseline for tasks in the music domain. In most cases of instrument classification, most LALMs, including GPT-4o, outperform Whisper-LLaMA. However, Whisper-LLaMA still outperforms most LALMs in tasks like ``Emotion Classification  In Songs'' and ``MARBLE Music Tagging'', with a few exceptions such that GPT-4o and Qwen-Audio-Chat beat LLaMA-Whisper in ``Emotion Classification In Songs'' and GAMA-IT and SALMONN-13B beat Whisper-LLaMA ``MARBLE Music Tagging''.  Despite its inaccessibility to acoustic information, Whisper-LLaMA relies solely on semantic content for predictions and still demonstrates superior performance compared to most LALMs. Besides, in tasks such as ``HEAR Percussion Instruments Classification'' and ``Instrument Source Classification'', Random baseline show comparable  performance with most LALMs.

In sum, the high refusal rates of GPT-4o in tasks within this domain poses a challenge in assessing its true capabilities for processing them. The best performance in ``Emotion Classification In Songs'', ``MARBLE Genre Classification'' and ``Music Genre Classification'' of GPT-4o among baselines suggests its potential to correctly identify the emotion and genres of the music.  However, from the tasks with lower refusal rate like ``MARBLE Music Tagging'', ``Instrument Classification'' and ``Instrument Source Classification'', GPT-4o fails to achieve the SOTA performance, reflecting its weakness in tagging music and classifying the instrument. These results suggest that GPT-4o is only comparable to other LALMs. Moreover, the comparable performance on some tasks between other LALMs and the two ineffective baselines, Whisper-LLaMA and Random, indicates that current LALMs struggle to capture and understand music-related acoustic information effectively for specific tasks.
\begin{table}[h!]
\centering
\begin{tabular}{p{0.47\textwidth}|p{0.48\textwidth}} 
\toprule
\midrule
\textbf{Task Name} & \textbf{Task Description} \\
\midrule
- Emotion Classification in Songs & Identify the domiant emotion of the music. \\
\midrule
- HEAR Music Genre Classification & \multirow{3}{0.45\textwidth}{Identify the genre of the music.} \\
- MARBLE Genre Classification & \\
- Music Genre Classification &  \\
\midrule 
- HEAR Percussion Instruments Classification & \multirow{2}{0.45\textwidth}{Identify the Beijing Opera percussion instrument appears in the music.} \\
- Instrument Classification - Beijing Opera & \\
\midrule
- HEAR Percussion Instruments Stroke Classification & Identify which stroke appears in the audio. \\
\midrule
- Instrument Classification - Nsynth & Identify which instrument appears in the audio. \\
\midrule
- Instrument Source Classification & Identify which type of instrument is present in the audio. \\
\midrule
- MARBLE Music Tagging & Identify the tag for the music. \\
\bottomrule
\end{tabular}
\vspace{10px}
\caption{Overview of tasks in the~\textbf{(Music) Music Classification} domain.}
\label{tab:tasks-audio-music-classification}
\end{table}

\clearpage

\subsection{Music Domain - Rhythm Analysis}

\begin{figure}[htbp]
    \centering
     \includegraphics[width=\textwidth]{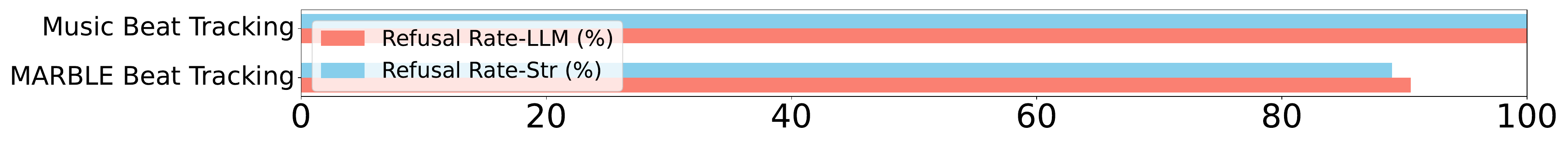} 
    \caption{Refusal rates of tasks in the~\textbf{(Music) Rhythm Analysis} domain. }
    \label{fig:rejection-music-rhythm-analysis}
    \end{figure}

The refusal rates of tasks about~\textbf{Rhythm Analysis} are demonstrated in Table~\ref{tab:tasks-music-rhythm-analysis}, Figure~\ref{fig:rejection-music-rhythm-analysis} respectively. As Figure~\ref{fig:rejection-music-rhythm-analysis} shows, GPT-4o refuses both tasks in this domain. Besides, no predictions can be extracted from responses produced by GPT-4o on ``MARBLE Beat Tracking'', resulting in a 100\% N/A rate. As these two tasks appear to be not associated with safety concerns outlined in the official report, we speculate that the reason GPT-4o refuses to respond is that the tasks are too hard for GPT-4o to complete accurately. The refusals are likely attributed to an effort to avoid generating misinformation.

\begin{table}[h!]
\centering
\begin{tabular}{p{0.4\textwidth}|p{0.55\textwidth}} 
\toprule
\midrule
\textbf{Task Name} & \textbf{Task Description} \\
\midrule
- Music Beat Tracking & \multirow{2}{0.55\textwidth}{Write down the timing of beats in the music.} \\
- MARBLE Beat Tracking & \\
\midrule 
\bottomrule
\end{tabular}
\vspace{10px}
\caption{Overview of tasks in the~\textbf{(Music) Rhythm Analysis} domain.}
\label{tab:tasks-music-rhythm-analysis}
\end{table}

\vspace{30px}
\subsection{Audio Domain - Quality Assessment}

\begin{figure}[h!]
    \centering
     \includegraphics[width=\textwidth]{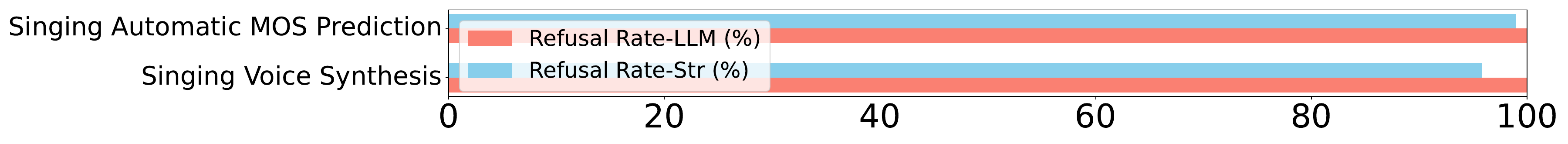} 
    \caption{Refusal rates of tasks in the~\textbf{(Audio) Quality Assessment} domain. }
    \label{fig:rejection-audio-quality-assessment}
    \end{figure}

The refusal rates of tasks about~\textbf{Quality Assessment} are demonstrated in Table~\ref{tab:tasks-audio-quality-assessment}, Figure~\ref{fig:rejection-audio-quality-assessment} respectively. In this case, the refusal rate determined by LLaMA-3.1-8B-Instruct is more accurate than string match methods, as GPT-4o sometimes responds with lyrics transcription rather than template-based refusals. That is, GPT-4o completely refused to perform both tasks in this domain. We speculate that the refusal of ``Singing Automatic MOS Prediction'' stems from concerns related to ungrounded inference and the refusal for ``Singing Voice Synthesis'' is attributed to the generation of copyrighted material. The former involves scoring the singer without any objective criteria, while the latter entails synthesizing a singing performance beyond the scope of the preset speaker.

\begin{table}[h!]
\centering
\begin{tabular}{p{0.38\textwidth}|p{0.55\textwidth}} 
\toprule
\midrule
\textbf{Task Name} & \textbf{Task Description} \\
\midrule
- Singing Automatic MOS Prediction & Write down the mean opinion score (MOS) of the audio. \\
\midrule
- Singing Voice Synthesis & Convert the given musical score into a singing audio.\\
\midrule 
\bottomrule
\end{tabular}
\vspace{10px}
\caption{Overview of tasks in the~\textbf{(Audio) Quality Assessment} domain.}
\label{tab:tasks-audio-quality-assessment}
\end{table}

\clearpage

\subsection{Audio Domain - Safety}

\begin{figure}[htbp]
    \centering
    \includegraphics[width=\textwidth]{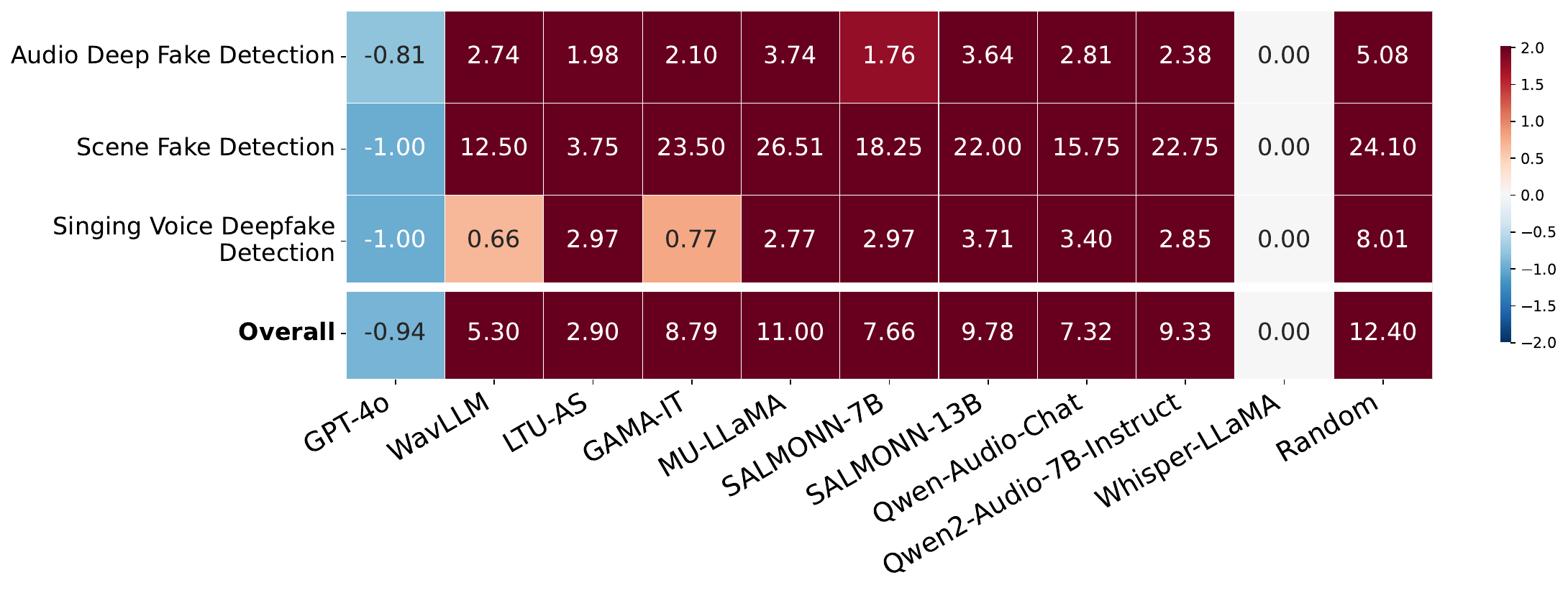} 
    \caption{Relative performance comparison of models in the~\textbf{(Audio) Safety} domain. }
    \label{fig:heatmap-audio-safety}
\end{figure}

\begin{figure}[htbp]
    \centering
     \includegraphics[width=\textwidth]{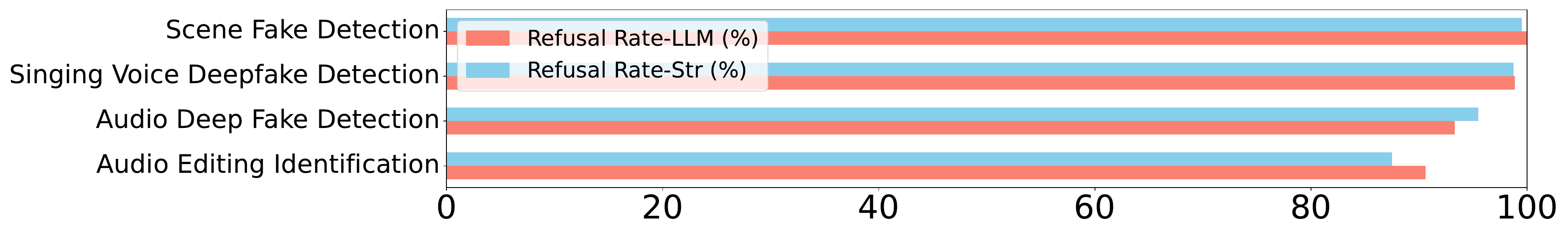} 
    \caption{Refusal rates of tasks in the~\textbf{(Audio) Safety} domain. }
    \label{fig:rejection-audio-safety}
    \end{figure}

The overview, relative scores, and refusal rates of tasks about~\textbf{Safety} are demonstrated in Table~\ref{tab:tasks-audio-safety}, Figure~\ref{fig:heatmap-audio-safety} and Figure~\ref{fig:rejection-audio-safety} respectively. Although the tasks in this domain do not appear to be associated with the safety concerns outlined in the official report, the notably high refusal rate for these tasks suggests that GPT-4o may have been post-trained to block requests related to deepfake detection.

Since detecting deepfake audio through text transcription is impossible, all LALMs, except GPT-4o, outperform Whisper-LLaMA. Due to its nearly complete refusal to respond,  GPT-4o achieves accuracies of only  1.57\%, 0\%, 0\% on ``Audio Deep Fake Detection'', ``Scene Fake Detection'' and ``Singing Voice Deepfake Detection'', respectively. Besides, the prediction cannot be extracted from all responses produced by GPT-4o on ``Audio Editing Identification'', resulting in a 100\% N/A rate.  Among other LALMs,  Random baseline outperforms these models, reflecting the accuracy of these LALMs are significantly lower than 50\% as these tasks are binary classification tasks. These results indicate substantial room for improvement in current LALMs.

\begin{table}[h!]
\centering
\begin{tabular}{p{0.35\textwidth}|p{0.6\textwidth}} 
\toprule
\midrule
\textbf{Task Name} & \textbf{Task Description} \\
\midrule
- Audio Deep Fake Detection &  Determine if the instrument sound is generated by machine. \\
\midrule
- Scene Fake Detection & Determine if the background noise is modified in the audio.\\
\midrule
- Singing Voice Deepfake Detection & Determine if the singing voice is generated by machine.\\
\midrule
- Audio Editing Identification & Write down the audio modifications, including instances of clipping, overlapping tracks, or any other edits \\
\midrule 
\bottomrule
\end{tabular}
\vspace{10px}
\caption{Overview of tasks in the~\textbf{(Audio) Safety} domain.}
\label{tab:tasks-audio-safety}
\end{table}

\clearpage


\subsection{Audio Domain - Signal-Characteristics Analysis}

\begin{figure}[htbp]
    \centering
    \includegraphics[width=\textwidth]{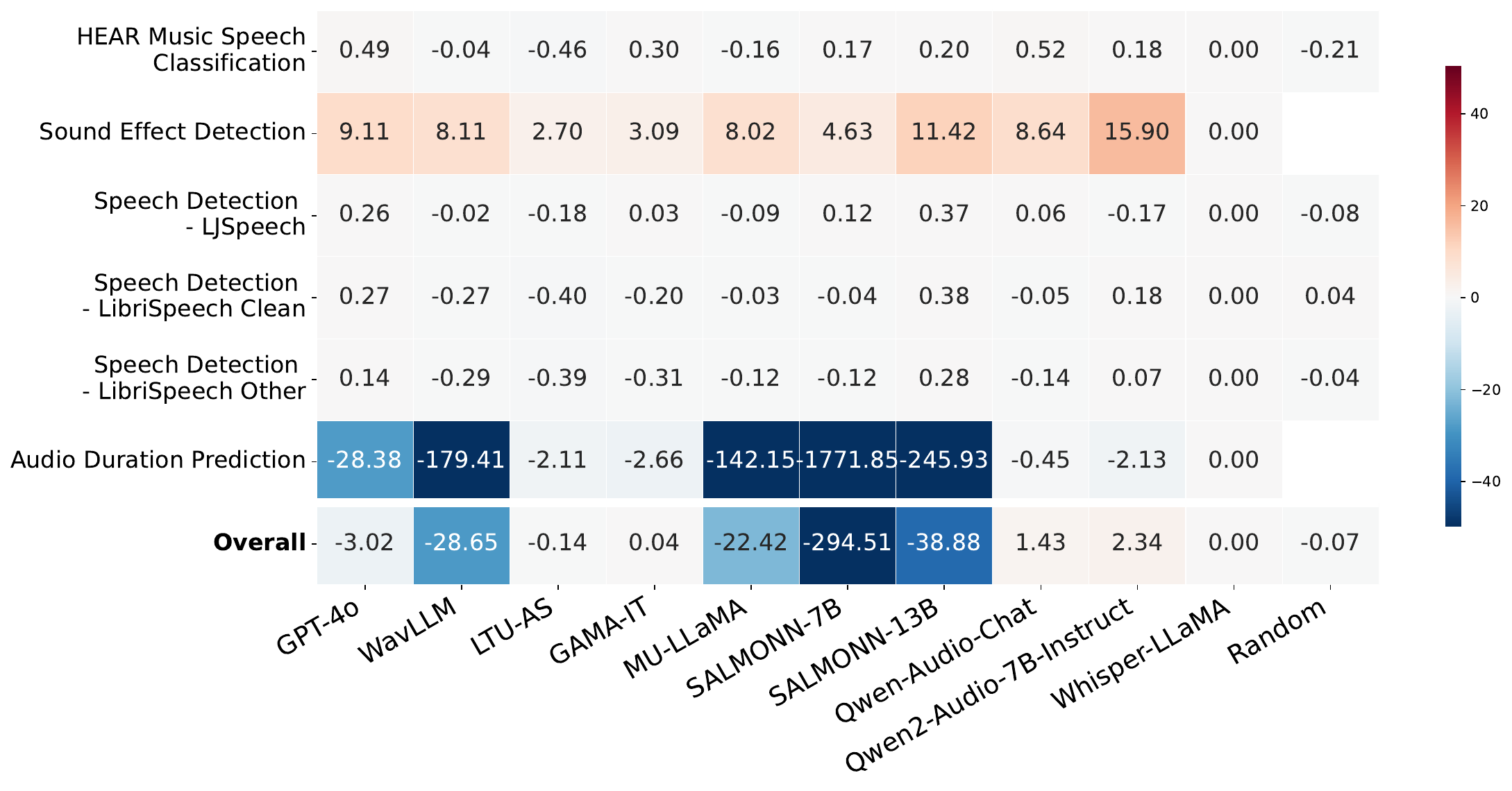} 
    \caption{Relative performance comparison of models in the~\textbf{(Audio) Signal-Characteristics Analysis} domain. }
    \label{fig:heatmap-audio-signal-characteristics-analysis}
\end{figure}

\begin{figure}[htbp]
    \centering
     \includegraphics[width=\textwidth]{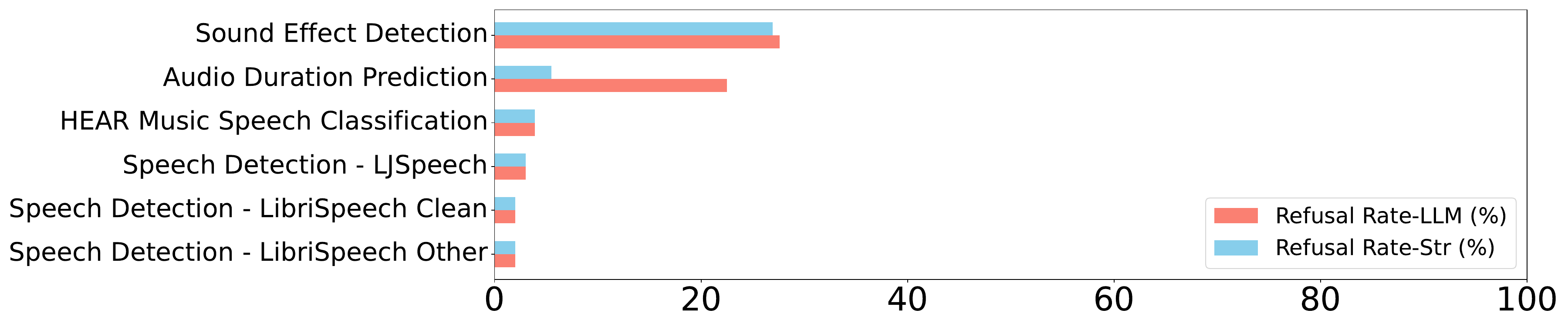} 
    \caption{Refusal rates of tasks in the~\textbf{(Audio) Signal-Characteristics Analysis} domain. }
    \label{fig:rejection-audio-signal-characteristics-analysis}
    \end{figure}

The overview, relative scores, and refusal rates of tasks about~\textbf{Signal-Characteristics Analysis} are demonstrated in Table~\ref{tab:tasks-audio-signal-characteristics-analysis}, Figure~\ref{fig:heatmap-audio-signal-characteristics-analysis} and Figure~\ref{fig:rejection-audio-signal-characteristics-analysis} respectively. Based on our inspection, no task in this domain is associated with safety concerns.

In tasks involving the detection of music, speech, or sound effects, GPT-4o outperforms most baselines. However, in  ``Audio Duration Prediction'', GPT-4o struggles to accurately estimate the duration of input audio. In contrast, other baselines, such as GAMA-IT, LTU-AS, Qwen-Audio-Chat, and Qwen2-Audio-7B-Instruct, predict duration effectively, indicating GPT-4o's limitations in this area.
 
\begin{table}[h!]
\centering
\begin{tabular}{p{0.4\textwidth}|p{0.55\textwidth}} 
\toprule
\midrule
\textbf{Task Name} & \textbf{Task Description} \\
\midrule
- HEAR Music Speech Classification & Determine the sound is speech or music. \\
\midrule
- Sound Effect Detection & Identify the specific audio effect in the audio. \\
\midrule
- Speech Detection & Determine if the given audio contains real speech. \\
\midrule
- Audio Duration Prediction & Write down the duration of the audio. \\
\midrule 
\bottomrule
\end{tabular}
\vspace{10px}
\caption{Overview of tasks in the~\textbf{(Audio) Signal-Characteristics Analysis} domain.}
\label{tab:tasks-audio-signal-characteristics-analysis}
\end{table}

\clearpage


\subsection{Audio Domain - Singing Analysis}

\begin{figure}[htbp]
    \centering
    \includegraphics[width=\textwidth]{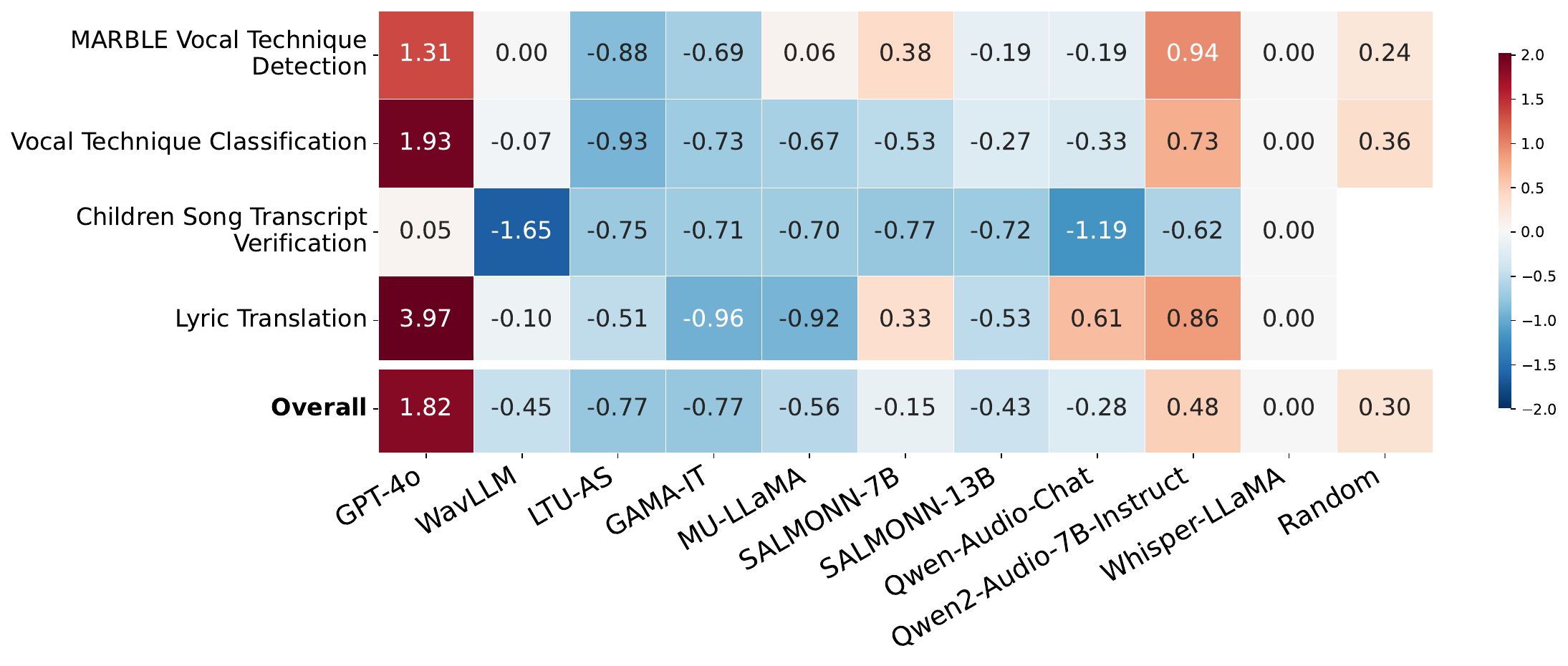} 
    \caption{Relative performance comparison of models in the~\textbf{(Audio) Singing Analysis } domain. }
    \label{fig:heatmap-audio-singing-analysis}
\end{figure}

\begin{figure}[htbp]
    \centering
     \includegraphics[width=\textwidth]{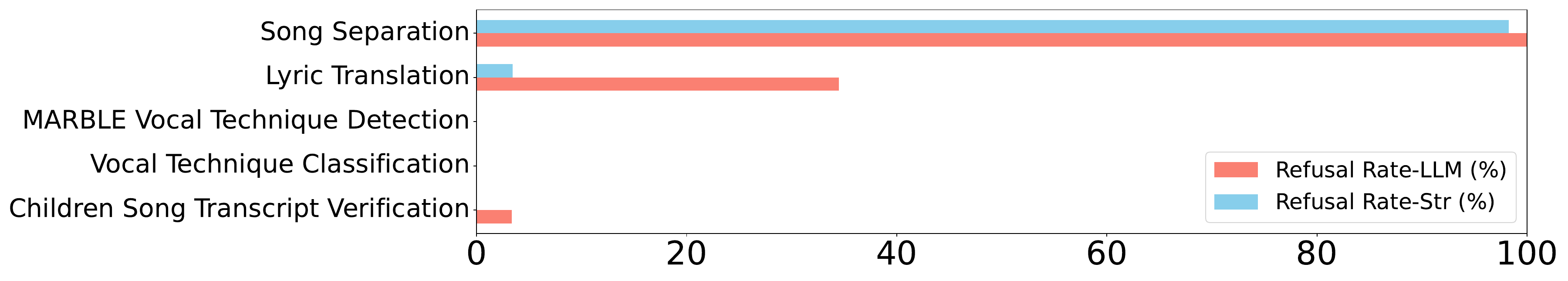} 
    \caption{Refusal rates of tasks in the~\textbf{(Audio) Singing Analysis } domain. }
    \label{fig:rejection-audio-singing-analysis}
    \end{figure}

The overview, relative scores, and refusal rates of tasks about~\textbf{Singing Analysis} are demonstrated in Table~\ref{tab:tasks-audio-singing-analysis}, Figure~\ref{fig:heatmap-audio-singing-analysis} and Figure~\ref{fig:rejection-audio-singing-analysis} respectively. ``Song Separation'' is a task aimed at separating piano music from vocals and other sounds in audio. We speculate that the task is related to unauthorized voice generation, as the singers in the songs are not the preset voices of GPT-4o. Beyond this task, GPT-4o tends to respond to other tasks. The high refusal rate of ``Lyric Translation'' determined by LLaMA-3.1-8B-Instruct appears to be a misjudgment, as it mistakenly classifies some correct outputs as refusals based on our manual inspection.

The vocal techniques conveyed in the music are lost in the transcription. Whisper-LLaMA relies solely on its pretrained knowledge and cannot interpret the music it hears. However, only GPT-4o and Qwen2-Audio-7B-Instruct outperform it and the Random baseline in related tasks, highlighting the limitations of many current LALMs in this area. For lyric translation, "Children's Song Transcript Verification" involves transcribing Korean songs into Korean words, while "Lyric Translation" requires both transcription and translation of the songs simultaneously. GPT-4o outperforms all baselines in these two tasks, with most baselines performing worse than Whisper-LLaMA, demonstrating GPT-4o's superior capabilities in singing analysis.
\begin{table}[h!]
\centering
\begin{tabular}{p{0.4\textwidth}|p{0.55\textwidth}} 
\toprule
\midrule
\textbf{Task Name} & \textbf{Task Description} \\
\midrule
- Song Separation & Separate the piano music from the vocals and other instruments in the given audio file. \\
\midrule
- MARBLE Vocal Technique Detection & \multirow{2}{0.455\textwidth}{Identify which vocal technique is used in the audio.}\\
- Vocal Technique Classification &  \\
\midrule
- Children Song Transcript Verification & Write down the lyrics in Korean.\\
\midrule
- Lyric Translation & Write down the lyrics in a specific language. \\
\midrule 
\bottomrule
\end{tabular}
\vspace{10px}
\caption{Overview of tasks in the~\textbf{(Audio) Singing Analysis} domain.}
\label{tab:tasks-audio-singing-analysis}
\end{table}

\vspace{-15px}

\clearpage


\subsection{Audio Domain - Sound Event}

\begin{figure}[htbp]
    \centering
    \includegraphics[width=\textwidth]{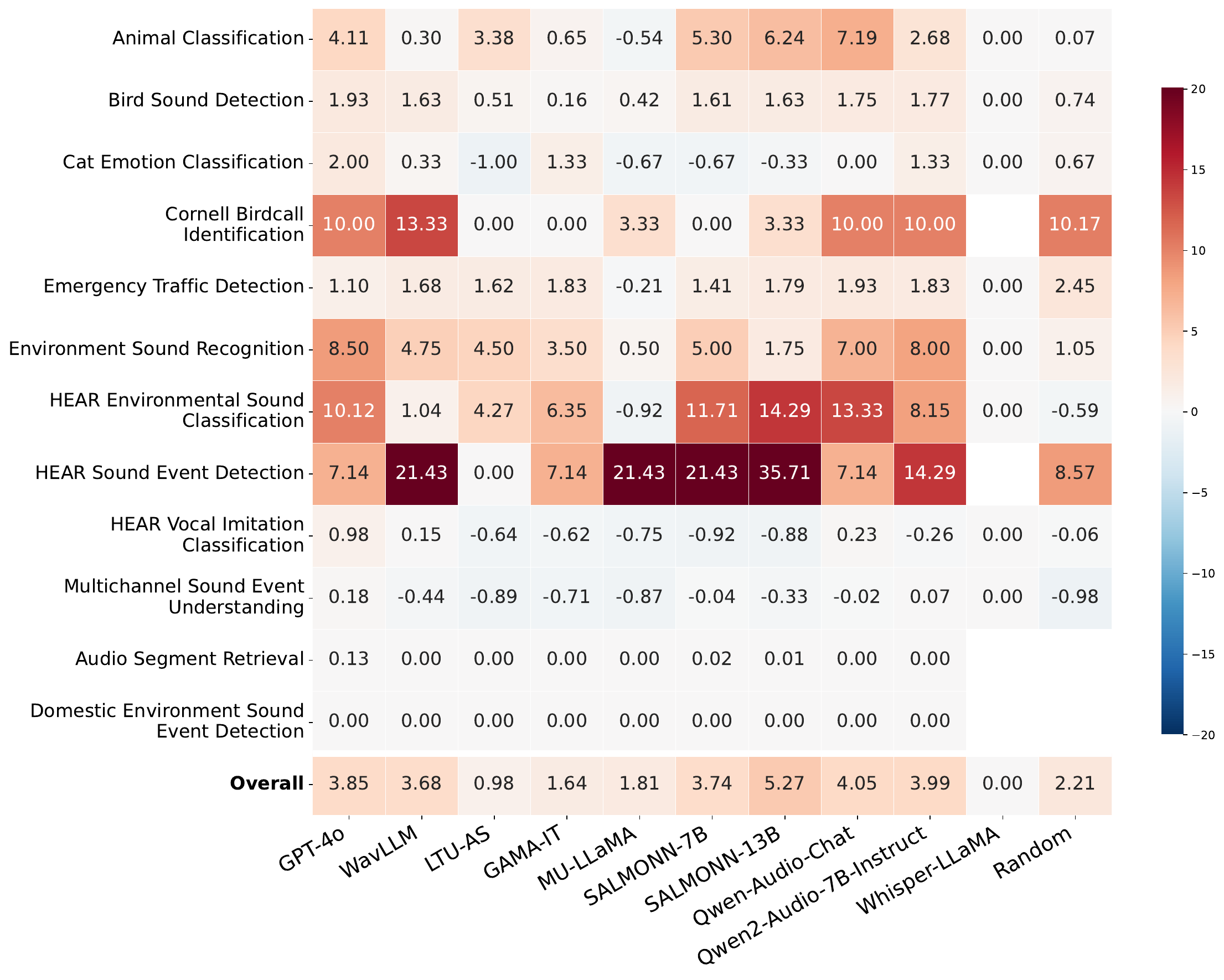} 
    \caption{Relative performance comparison of models in the~\textbf{(Audio) Sound Event} domain. }
    \label{fig:heatmap-audio-sound-event}
\end{figure}

\begin{figure}[htbp]
    \centering
     \includegraphics[width=\textwidth]{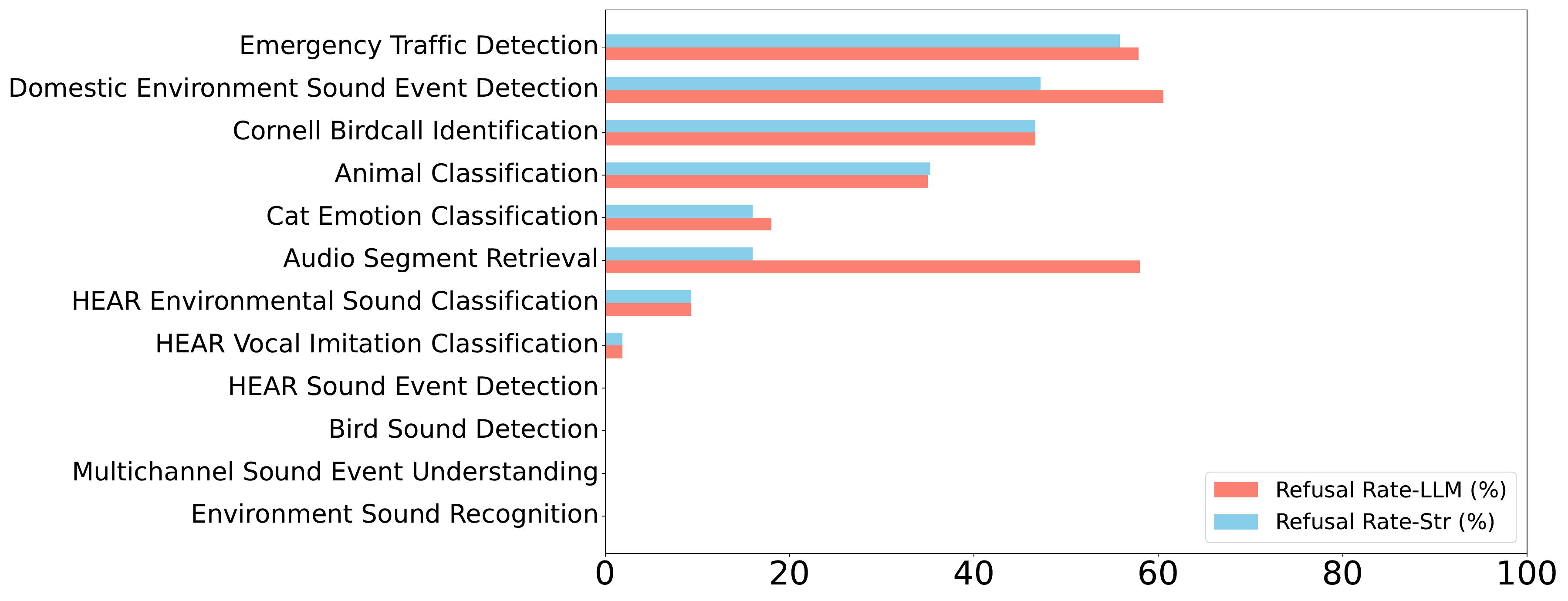} 
    \caption{Refusal rates of tasks in the~\textbf{(Audio) Sound Event} domain. }
    \label{fig:rejection-audio-sound-event}
    \end{figure}

The overview, relative scores, and refusal rates of tasks about~\textbf{Sound Event} are demonstrated in Table~\ref{tab:tasks-audio-sound-event}, Figure~\ref{fig:heatmap-audio-sound-event} and Figure~\ref{fig:rejection-audio-sound-event} respectively. Based on our inspection, no tasks in this domain are associated with safety issues. Hence, we speculate that the high refusal rate of tasks in this domain is due to GPT-4o's low confidence in processing these tasks, leading it to refuse some samples to avoid misinformation. Additionally, LLaMA-3.1-8B-Instruct predicts an incorrect refusal rate for ``Audio Segment Retrieval''.  According to our manual inspection, the refusal rate determined by string matching is more accurate.

As it's challenging for models to solves the tasks in this domain by the transcription produced by Whisper, most LALMs including GPT-4o outperform Whisper-LLaMA in this domain. Also, they beat Random baselines on most tasks except for ``Emergency Traffic Detection'', a binary classification problem. For animal and environmental sound classification tasks, Qwen-Audio-Chat and SALMONN slightly outperform GPT-4o. Interestingly, while GPT-4o's performance in these tasks is only average, it achieves significant improvement in 'HEAR Vocal Imitation Classification', a task involving human-mimicked sound events, compared to other LALMs.  In ``Multichannel Sound Event Understanding'', GPT-4o outperforms other LALMs and is the only two LALMs beats Whisper-LLaMA, demonstrating its ability to capture spatial and temporal information effectively.

Despite the promising relative score of GPT-4o shown in~Figure~\ref{fig:heatmap-audio-sound-event}, there is still room for improvement of the current LALMs. In ``Multichannel Sound Event Understanding'', GPT-4o achieves the highest accuracy, but it is only 45.77\%. In ``Audio Segment Retrieval'' and ``Domestic Environment Sound Event Detection'', GPT-4o achieves a 0.1336 segment IOU and a 0.00079 event-based F1 score, with most baselines getting a score of 0. These results indicate that such complex tasks remain challenging for current LALMs.

\begin{table}[h!]
\centering
\begin{tabular}{p{0.5\textwidth}|p{0.45\textwidth}} 
\toprule
\midrule
\textbf{Task Name} & \textbf{Task Description} \\
\midrule
- Animal
Classification & Identify which animal voice presents in the audio. \\
\midrule 
- Bird Sound Detection & Determine if any bird vocalizations in the speech. \\
\midrule 
- Cat Emotion Classification & Identify the cat's emotion from the audio. \\
\midrule 
- Cornell Birdcall Identification &  Identify the species of bird whose voice is present in the audio. \\
\midrule 
- Emergency Traffic Detection & Identify what's the  emergency traffic alerts. \\ 
\midrule 
- Environment Sound Recognition & \multirow{3}{0.45\textwidth}{Identify which environmental sound presents in the audio.} \\
- HEAR Environmental Sound Classification &  \\
- Domestic Environment Sound Event Detection &  \\
\midrule 
- HEAR Sound Event Detection & Identify the type of the sound. \\
\midrule 
- HEAR Vocal Imitation Classification & Identify the type of the vocal imitation. \\
\midrule 
- Multichannel Sound Event Understanding & Listen to the given multichannel audio and answer a multiple-choice question about the sound event, distance, angle, or timestamps.  \\
\midrule 
- Audio Segment Retrieval & Write down the starting and stopping timestamps for the specified audio section.\\

\midrule 
\bottomrule
\end{tabular}
\vspace{10px}
\caption{Overview of tasks in the~\textbf{(Audio) Sound Event} domain.}
\label{tab:tasks-audio-sound-event}
\end{table}

\clearpage


\subsection{Audio Domain - Spatial Audio Analysis}

\begin{figure}[htbp]
    \centering
    \includegraphics[width=\textwidth]{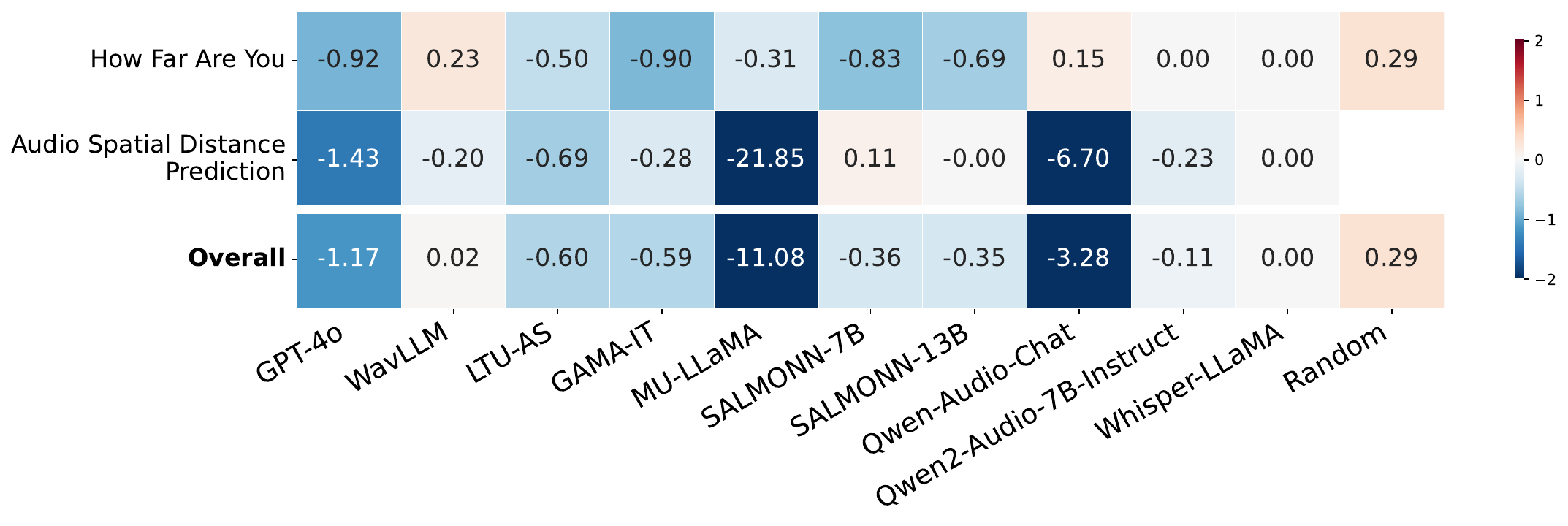} 
    \caption{Relative performance comparison of models in the~\textbf{(Audio) Spatial Audio Analysis} domain. }
    \label{fig:heatmap-audio-spatial-audio-analysis}
\end{figure}

\begin{figure}[htbp]
    \centering
     \includegraphics[width=\textwidth]{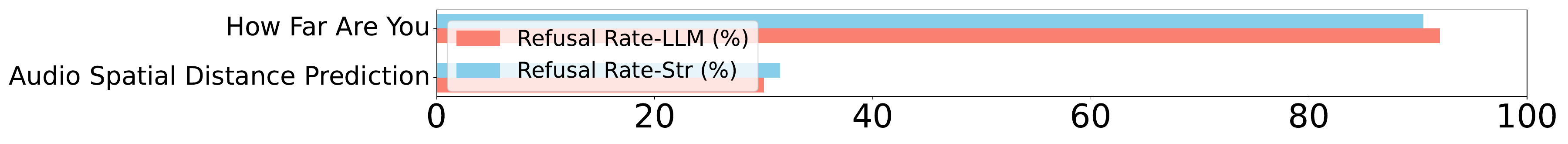} 
     
    \caption{Refusal rates of tasks in the~\textbf{(Audio) Spatial Audio Analysis} domain. }
    \label{fig:rejection-audio-spatial-audio-analysis}
    \end{figure}

The overview, relative scores, and refusal rates of tasks about~\textbf{Spatial Audio Analysis} are demonstrated in Table~\ref{tab:tasks-audio-spatial-audio-analysis}, Figure~\ref{fig:heatmap-audio-spatial-audio-analysis} and Figure~\ref{fig:rejection-audio-spatial-audio-analysis} respectively. GPT-4o refuses to respond to most samples in the ``How Far Are You'' task, while still selecting an answer for approximately 10\% of the samples based on the instructions provided. For the open-ended regression task "Audio Spatial Distance Prediction," GPT-4o demonstrates a significantly lower refusal rate. In this case, GPT-4o often provides a response with an estimated range, such as ``The distance is approximately 1.5 to 2 meters.'' By comparing these two tasks, we speculate that GPT-4o's refusals may stem from its avoidance of misinformation, as determining a precise distance is inherently challenging. However, providing an approximate range seems to be more feasible, leading to a higher response rate.

The spatial features of audio are completely lost when the speech is transcribed into text, making the cascaded system essentially make random guesses. However, most LALMs perform significantly worse than Whisper-LLaMA. In the classification task ``How Far Are You'', even a Random baseline outperforms other models. These results highlight that current LALMs are unable to effectively capture spatial information or respond accurately to related questions.

\begin{table}[h!]
\centering
\begin{tabular}{p{0.35\textwidth}|p{0.6\textwidth}} 
\toprule
\midrule
\textbf{Task Name} & \textbf{Task Description} \\
\midrule
- How Far Are You & Identify how far the speaker is from the microphone. \\
\midrule
- Audio Spatial Distance Prediction & Write down the spatial distance based on the given audio. \\
\midrule 
\bottomrule
\end{tabular}
\vspace{10px}
\caption{Overview of tasks in the~\textbf{(Audio) Spatial Audio Analysis} domain.}
\label{tab:tasks-audio-spatial-audio-analysis}
\end{table}

\clearpage

\section{MMAU}
\label{sec:mmau}
Massive Multi-Task Audio Understanding and Reasoning Benchmark (MMAU)~\cite{sakshi2024mmau} is designed to assess the LMMs' reasoning and understanding abilities. Each sample in MMAU is comprised of an expert-written question and then formulated into a multi-choice problem. The benchmark spans the Audio, Music, and Speech domains, including the assessment of capabilities like Phonemic Stress Pattern Analysis, Conversational Fact Retrieval, Key Highlight Extraction, and others. MMAU are designed in the form of two phases: Test-mini and Test, with a percentage of 1:9. The ground truth of the former is public-available, and that of the latter is close-sourced. 

Thus, to probe the owning of expert knowledge of GPT-4o, we conduct an evaluation on MMAU. As shown in Table~\ref{tab:mmau-result}, GPT-4o outperforms existing open-source audio-language models across various domains and surpasses the proprietary model Gemini Pro v1.5~\cite{team2024gemini}. Notably, GPT-4o beats the cascaded pipelines in the Sound and Music domains. Note that the cascaded pipelines adopted here are the combination of an audio caption model and a textual LLM like GPT-4o or LLaMA-3-8B-Instruct. The audio caption model adopted here is Qwen-2-Audio-Instruct. 

Among the 10,000 samples in MMAU, the refusal rate, as determined by string matching and LLM judge, is 0.76\% and 1.38\%, respectively.

\begin{table}[!h]
\centering

\resizebox{\columnwidth}{!}{
    \begin{tabular}{lcccccccc}
    \toprule
    \multirow{2}{*}{\textbf{Models}} & \multicolumn{2}{c}{\textbf{Sound}} & \multicolumn{2}{c}{\textbf{Music}} & \multicolumn{2}{c}{\textbf{Speech}} & \multicolumn{2}{c}{\textbf{Avg}} \\
    \cmidrule(lr){2-3} \cmidrule(lr){4-5} \cmidrule(lr){6-7} \cmidrule(lr){8-9}
     & \textbf{Test-mini} & \textbf{Test} & \textbf{Test-mini} & \textbf{Test} & \textbf{Test-mini} & \textbf{Test} & \textbf{Test-mini} & \textbf{Test} \\
    \midrule
    Random Guess & 26.72 & 25.73 & 24.55 & 26.53 & 25.50 & 26.72 & 26.00 & 25.92 \\
    Most Frequent Choice & 27.02 & 25.73 & 20.35 & 23.73 & 29.12 & 30.33 & 25.50 & 26.50 \\
    Human & 86.31 & - & 78.22 & - & 82.17 & - & 82.23 & - \\
    \midrule
    LTU-AS & 23.35 &  24.96 &  9.10 & 10.46  & 20.60 & 21.30 & 17.68 & 18.90 \\
    MU-LLaMA & 40.84 & 44.80 & 32.63 & 30.63 & 22.22 & 16.56 & 31.90 & 30.66 \\
    GAMA-IT & 43.24 & 43.23 & 28.44 & 28.00 & 18.91 & 15.84 & 30.20 & 29.02 \\
    Qwen-Audio-Chat & 55.25 & 56.73 & 44.00 & 40.90 & 30.03 & 27.95 & 43.10 & 41.86 \\
    Qwen2-Audio-7B-Instruct & 54.95 & 45.90 & 50.98 & 53.26 & 42.04 & 45.90 & 49.20 & 52.50 \\
    SALMONN-13B & 41.00 & 40.30 & 34.80 & 33.76 & 25.50 & 24.24 & 33.70 & 32.77 \\
    \midrule
    Gemini Pro v1.5 & 56.75 & 54.46 & 49.40 & 48.56 & \textbf{58.55} & 55.90 & 54.90 & 52.97 \\
    \midrule
    GPT-4o Voice Mode & \textbf{63.36} & \textbf{62.83} & \textbf{60.77} & \textbf{54.73} & 53.15 & \textbf{63.80} & \textbf{59.10} & \textbf{60.46} \\
    \midrule
    (cascade) GPT-4o  & 57.35 & 55.83 & 49.70 & 51.73 & 64.66 & 68.66 & 57.30 & 58.74 \\
    (cascade) LLaMA-3-Instruct& 50.75 & 49.10 & 50.29 & 48.93 & 55.25 & 62.70 & 52.10 & 53.57 \\
    
    \bottomrule
    \end{tabular}
}
\vspace{1ex}
\caption{MMAU results}
\label{tab:mmau-result}
\end{table}

\section{CMM}
\label{sec:cmm}
CMM~\cite{leng2024curse} aims to gauge the hallucination level of Large Multi-modal Models (LMMs). While LMMs can process multi-modal inputs simultaneously, their ability to accurately interpret relationships across different modalities is still uncertain. CMM introduces a systematic test covering text, audio, and visual samples to evaluate models' reliance on unimodal biases and unintended inter-modality associations. 

In our analysis, we focus solely on audio-text correlations, as the version of GPT-4o we used is limited to audio-language inputs, making it infeasible to handle visual data. The audio-language tests are designed to detect spurious correlations arising from global appearance frequencies and co-occurrence frequencies. The former involves objects or events commonly appearing in existing datasets. They evaluate whether higher frequency in the training data causes LMMs to hallucinate their presence, even when they are absent in the input. The latter collects pairs of objects that frequently co-occur during training and checks whether model incorrectly predicts the presence of one object when only the other is present.

The two core metrics used in CMM is Perception Accuracy (PA) and Hallucination Resistance (HR): 

\begin{equation}
\text{PA} = \frac{\# \text{correctly predicted ``yes''}}{\# \text{ground truth ``yes''}} \quad \text{HR} = \frac{\# \text{correctly predicted ``no''}}{\# \text{ground truth ``no''}}
\end{equation}

\begin{table}[!ht]
\centering
\begin{tabular}{lrr}
\midrule
        & PA $\uparrow$ & HR $\uparrow$ \\
    \midrule
    Qwen2-Audio      & \textbf{98.50} \% & 34.50 \%   \\
    Audio-Flamingo   & 89.50 \% & 39.00 \%  \\
    GAMA-IT       & 94.50 \% & 52.00 \%  \\
    SALMONN   &  93.00 \% & 59.00 \%  \\
    \midrule
    GPT-4o   & 89.00 \% & \textbf{83.75} \% \\
\midrule 
\end{tabular}
\vspace{5px}
\caption{CMM result.}
\label{tab:cmm-result}
\end{table}

Based on the evaluation results in Table~\ref{tab:cmm-result}, GPT-4o appears to be less sensitive to detecting actual events, with a slightly lower PA compared to other models. However, its HR is significantly higher, indicating much stronger resistance to hallucinating sounds that are not present.

Among the 400 samples in CMM, the refusal rates determined by both string matching and LLM evaluation are 1\%.

\section{Conclusions}
In this report, we present a comprehensive evaluation of GPT-4o’s audio understanding and reasoning capabilities through extensive experiments across a diverse range of tasks. By analyzing the model’s performance based on multiple criteria, we assess its effectiveness on large-scale benchmarks, including Dynamic-SUPERB , MMAU, and CMM, spanning the domains of audio, speech, and music. These benchmarks allow us to examine GPT-4o’s ability to interpret acoustic information from audio inputs and generate responses based on relevant observations.

Due to the significant privacy and ethical concerns surrounding the abundant information carried by audio, it is crucial to ensure the responsible use of voice assistant systems. GPT-4o demonstrates superior audio understanding and reasoning capabilities and serves as a pioneering end-to-end spoken language model that incorporates post-training to enhance safety. However, as demonstrated through our experiments and analysis, the current version remains highly sensitive to input prompts, occasionally refusing to process tasks that do not pose any safety risks. Striking a balance between ethical considerations and advanced audio comprehension is significantly more complex than in the text domain, as acoustic features carry richer and more nuanced information than semantic content alone. Addressing these challenges remains a challenge for the speech research community.

\section*{Acknowledgment}
We sincerely appreciate all the contributors involved in the development of LALMs and benchmarks, whose efforts have provided us with valuable insights, tools, and resources that have greatly supported this report. We also acknowledge that model performance may vary depending on the evaluation protocols used. As such, this report serves as an initial exploration of the current state of LALMs, and we look forward to further refinements and discussions in the future.

{
\bibliographystyle{plain}
\bibliography{main}
}

\end{document}